\documentclass[10pt,twocolumn,letterpaper]{article}

\usepackage{algorithm}
\usepackage{algorithmicx}
\usepackage{algpseudocode}
\usepackage{multirow}
\usepackage{amssymb}
\usepackage{pifont}
\usepackage{makecell}
\usepackage{tikz}
\usepackage{float}
\usepackage{stfloats}
\usepackage{graphicx}
\usepackage{afterpage}

\usepackage{listings}
\usepackage{xcolor}
\usepackage[T1]{fontenc}  % <--- 添加这一行

\usepackage{iccv}

\definecolor{iccvblue}{rgb}{0.21,0.49,0.74}
\usepackage[pagebackref,breaklinks,colorlinks,citecolor=iccvblue]{hyperref}

\def\paperID{11314} % *** Enter the Paper ID here
\def\confName{ICCV}
\def\confYear{2025}

\title{Enhancing Adversarial Transferability by Balancing Exploration and Exploitation with Gradient-Guided Sampling}

\author{
    Zenghao Niu$^{1}$ \quad
    Weicheng Xie$^{1,2,3}$ \thanks{Corresponding author.} \quad
    Siyang Song$^{4}$ \quad
    Zitong Yu$^{5}$ \quad
    Feng Liu$^{1}$ \quad
    Linlin Shen$^{6}$ \quad \\
    $^{1}$School of Computer Science \& Software Engineering, Shenzhen University, China \\
    $^{2}$Guangdong Laboratory of Artificial Intelligence and Digital Economy (SZ), Shenzhen, China \\
    $^{3}$Guangdong Provincial Key Laboratory of Intelligent Information Processing, Shenzhen University, China \\
    $^{4}$School of Computer Science, University of Exeter, U.K. \\
    $^{5}$Department of Computing and Information Technology, Great Bay University, China \\
    $^{6}$Computer Vision Institute, School of Artificial Intelligence, Shenzhen University, China \\
    {\tt\small \{2300271072@mail., wcxie@, feng.liu@, llshen@\}szu.edu.cn } \\
    {\tt\small s.song@exeter.ac.uk, zitong.yu@ieee.org }
}

\begin{document}
\maketitle

\begin{abstract}
\vspace{-2em}

Adversarial attacks present a critical challenge to deep neural networks' robustness, particularly in transfer scenarios across different model architectures. However, the transferability of adversarial attacks faces a fundamental dilemma between Exploitation (maximizing attack potency) and Exploration (enhancing cross-model generalization). Traditional momentum-based methods over-prioritize Exploitation, i.e., higher loss maxima for attack potency but weakened generalization (narrow loss surface). Conversely, recent methods with inner-iteration sampling over-prioritize Exploration, i.e., flatter loss surfaces for cross-model generalization but weakened attack potency (suboptimal local maxima). To resolve this dilemma, we propose a simple yet effective Gradient-Guided Sampling (GGS), which harmonizes both objectives through guiding sampling along the gradient ascent direction to improve both sampling efficiency and stability. Specifically, based on MI-FGSM, GGS introduces inner-iteration random sampling and guides the sampling direction using the gradient from the previous inner-iteration (the sampling's magnitude is determined by a random distribution). This mechanism encourages adversarial examples to reside in balanced regions with both flatness for cross-model generalization and higher local maxima for strong attack potency. Comprehensive experiments across multiple DNN architectures and multimodal large language models (MLLMs) demonstrate the superiority of our method over state-of-the-art transfer attacks. Code is made available at https://github.com/anuin-cat/GGS. 
\end{abstract}
\section{Introduction}
\label{sec:intro}

\begin{figure}[!tb]
    \centering
    \includegraphics[width=0.48\textwidth ]{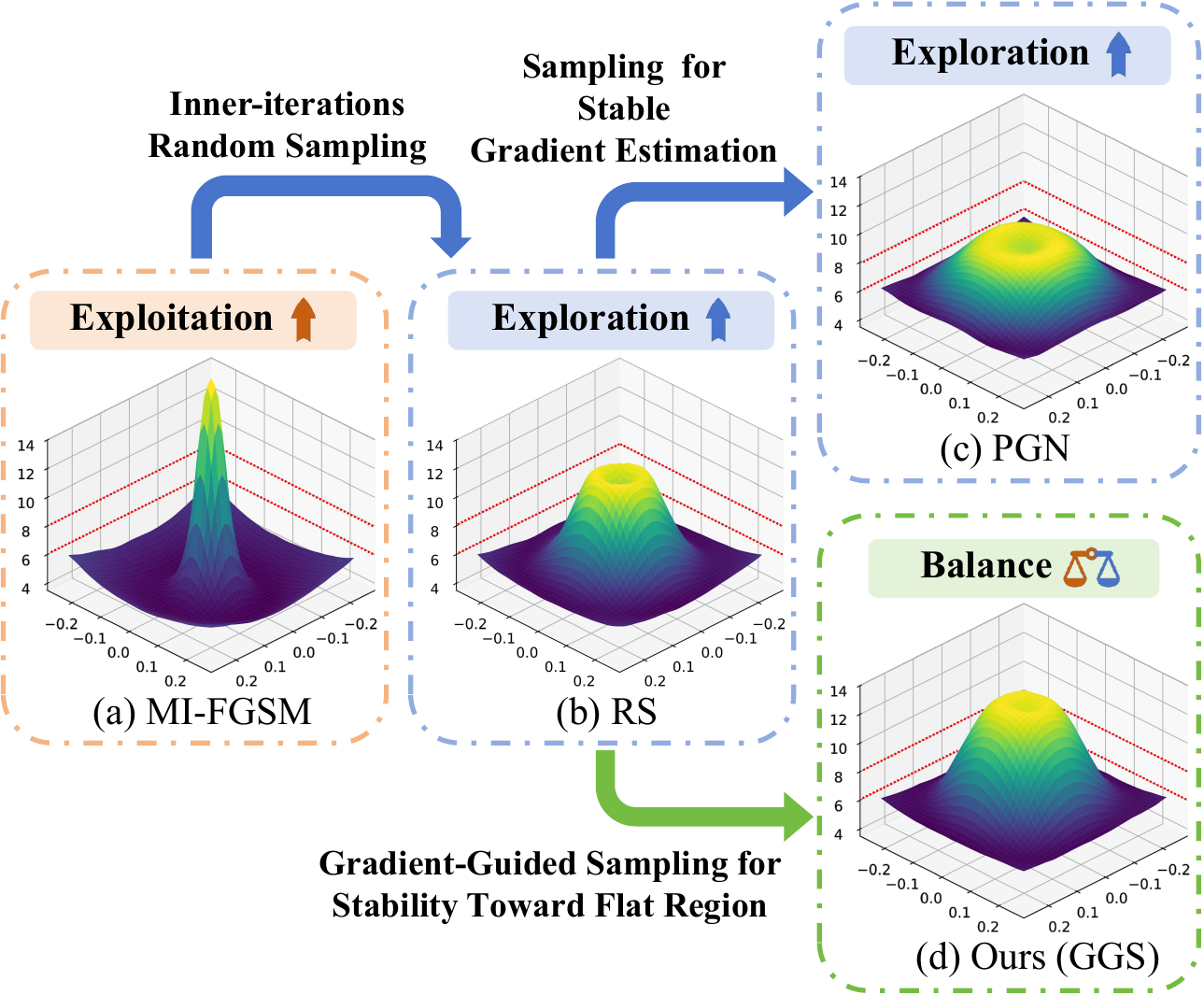}
    \caption{The loss surfaces of (a) MI-FGSM \cite{dong2018mifgsm} (Momentum iterative fast gradient sign method), (b) RS (Base inner-iteration Random Sampling defined in section \ref{motivation}) for enhancing exploration within the neighborhood, (c) PGN \cite{ge2023pgn} (Penalizing Gradient Norm) with Random Sampling (RS) for stable gradient estimation to enhance exploration, and 
    (d) Our GGS (Gradient-Guided Sampling) for efficient sampling to generate gradients that stably towards flat regions with higher local maxima.
    It shows that, building upon RS, our approach not only maintains a flat loss surface,  
    but also improves the local maximum loss value, compared to PGN, achieving a balance between exploration and exploitation.
    }
    \label{fig:first_img}
\end{figure}

Researches have indicated that adding imperceptible perturbations to input data can easily affect deep neural networks (DNNs), leading to erroneous predictions in safety-critical domains such as autonomous driving \cite{wu2023autodrive1, hossain2023autodrive3}, and cybersecurity \cite{ten2010cybersecurity1, husak2018cybersecurity2}.
To facilitate the development of relevant defense mechanisms, in-depth research on adversarial attack methods is urgently needed.

While adversarial attacks are categorized into white-box and black-box settings, the black-box scenario is more realistic in practical applications, where the attacker lacks knowledge of the target model
and relies solely on a surrogate model, aiming to generate adversarial examples that are effective across multiple target models. To enhance the transferability of adversarial examples in this setting, researchers have explored gradient-based methods \cite{dong2018mifgsm, wang2021vmifgsm, ge2023pgn}, input transformation methods \cite{dong2019tim, xie2019dim, wang2021admix}, and generative methods \cite{xiao2018advgan, liu2024advdiffusion, kang2023diffattack} to mitigate overfitting of adversarial examples on surrogate models. 
Despite significant progress, the effectiveness of attacks on target models remains substantially lower than that on surrogate models, driving the need for more effective attack strategies.

In black-box adversarial attacks, RAP \cite{qin2022rap} adopted Sharpness-Aware Minimization (SAM \cite{foret2020sam}) to improve transferability by flattening the loss surface.
Building on this, PGN \cite{ge2023pgn} further enhanced the transferability of adversarial examples by 
penalizing gradient norm and flattening the loss surface.
The efficacy of flat loss surface in improving adversarial attack transferability is increasingly supported \cite{ge2023pgn, fang2023elf, xu2023app}.
To enhance exploration through local flatness, recent approaches, building on MI-FGSM's \cite{dong2018mifgsm} momentum-based \textbf{outer-iterations}, often incorporates \textbf{inner-iterations} before each example update \cite{zhu2023gra, ge2023pgn}, i.e., performing multi-step neighborhood sampling for stabilization.

However, these methods enhance Exploration (cross-model generalization) at the cost of compromising Exploitation (attack potency).
This is because, although inner-iterations can significantly enhance exploration capability through neighborhood sampling, independent random sampling struggles to obtain stable and effective gradient directions, while potential noise may cause the averaged gradient to deviate from the optimal ascent toward target regions, which exhibit flatter loss surface and higher local maxima.
To optimize sampling efficiency, a guided sampling constraint is essential, which should satisfy two criteria: (1) ensuring stable gradient ascent {direction and (2) directing gradients toward flatter regions, thereby enabling samples to reach flatter local maxima regions with enhanced transferability, balancing exploitation and exploration.

For (1) stable gradient direction, a sufficiently small neighborhood sampling followed by gradient calculation generally yields correct ascent directions. Yet, how can we further stabilize this direction?
The Nesterov Accelerated Gradient (NAG) \cite{nesterov1983nag} enhances stability by firstly updating samples along the current momentum (lookahead) before acquiring gradients, providing more stable directions. 
Therefore, momentum-based direction constraints are promising for sampling, as the additional random magnitude can be used to preserve the sampling randomness.
(2) However, directly applying NAG's momentum mechanism during inner-iteration sampling imposes strong constraints on subsequent directions, restricts the exploration capability for flat regions. 
To address this, we replace momentum with the gradient from the previous inner-iteration, named Gradient-Guided Sampling (GGS). 
This adjustment ensures that sampling to align with gradient ascent directions while reducing the impact from early-stage instability. 
Our GGS can achieve stable sampling directions after brief oscillations, consistently targeting flat loss regions with larger local maxima regardless of the initial sampling quality.

As shown in Fig. \ref{fig:first_img}(d), our GGS generates adversarial examples with flatter loss surface and higher local maxima compared to RS, i.e., it does not sacrifice local maxima in the pursuit of a flatter surface compared to PGN, thereby achieving a balance between exploration and exploitation to facilitate attack transferability.
Additionally, since our approach represents a complementary improvement over Random Sampling (RS), it is compatible with RS-based methods, such as inner-iteration random sampling and gradient averaging, including PGN \cite{ge2023pgn} and GRA \cite{zhu2023gra}. In summary, this study makes the following contributions:

\begin{itemize}
    \item We propose Gradient-Guided Sampling (GGS), a novel inner-iteration sampling strategy that effectively balances exploration (cross-model generalization) and exploitation (attack potency). By leveraging the gradient from the previous inner-iteration to guide sampling directions, GGS achieves stable gradient ascent toward flat loss regions with higher local maxima.
    \item The GGS framework is compatible with existing inner-iteration random sampling-based methods, enhancing their sampling efficiency and further improving the transferability of the adversarial examples they generate.
    \item Extensive experimental results as well as comprehensive visualizations demonstrate the efficacy of our GGS, covering targeted and non-targeted attacks in cross-architecture black-box scenarios, non-targeted attacks on multimodal large language models (MLLMs) and commercial cloud functions.
\end{itemize}
\section{Related Work}
\label{sec:related_work}

Transferability is a critical property in adversarial attacks, allowing adversarial examples generated from a surrogate model to deceive unseen target models in black-box settings.
This characteristic makes transfer-based attacks highly practical in real-world applications where target models are inaccessible, motivating our focus on transferable adversarial attacks. 

\subsection{Transferable Attack}
\textbf{Gradient-based Attack.} 
Gradient-based attacks aim to generate adversarial examples by exploiting the gradients of a surrogate model to maximize the loss. Classic methods such as Fast Gradient Sign Method (FGSM) \cite{Goodfellow2015fgsm} and Projected Gradient Descent (PGD) \cite{madry2017pgd} optimize adversarial perturbations to fool the surrogate model. However, these attacks often overfit the surrogate model, resulting in limited transferability to target models. To address this, advanced techniques such as momentum-based (MI-FGSM \cite{dong2018mifgsm}), Nesterov Iterative-based (NI-FGSM \cite{lin2019nifgsm_sim}), variance-tuned (VMI-FGSM \cite{wang2021vmifgsm}), gradient relevance-based (GRA \cite{zhu2023gra}), momentum initialization (GI-FGSM \cite{wang2024gifgsm}) and distribution-based methods (ANDA \cite{fang2024anda}) have been introduced to smooth out the optimization process and improve transferability across models. 

\textbf{Input Transformation.}
To mitigate the overfitting issue inherent in gradient-based methods, input transformation techniques have been proposed to enhance the diversity of adversarial examples. Techniques like translation, resizing, padding, image mixup (e.g., DIM \cite{xie2019dim}, SIM \cite{lin2019nifgsm_sim}, TIM \cite{dong2019tim}, Admix \cite{wang2021admix}) and other advanced image transformation methods (e.g., SSM \cite{long2022ssm}, SIA \cite{wang2023sia}, STM \cite{ge2023stm}, L2T \cite{zhu2024l2t}, BSR \cite{wang2024bsr}) introduce randomness to the input data before generating perturbations. 
By applying such transformations during the attack process, more effective transferable adversarial examples can be generated. 

\subsection{Flat Maxima} 

The generalization ability of models has been suggested to possibly have a certain association with flat minima \cite{hochreiter1997flat1}. More in-depth research and exploration have gradually confirmed this perspective \cite{keskar2016flat2, li2018flat4, foret2020sam, zhao2022flat3}. Moreover, flat maxima also have been validated to be effective for enhancing the generalization and transferability of adversarial examples \cite{qin2022rap}. Flat maxima refers to regions in the loss landscape, where small changes in the model parameters result in minimal changes in the loss, making adversarial examples generated in these regions less sensitive to the specific decision boundaries of models. This property contrasts with ``sharp maxima'', which are highly sensitive to small perturbations and can lead to overfitting to the surrogate model, reducing transferability capacity.

In the context of transfer-based attacks, works like Reverse Adversarial Perturbation (RAP) \cite{qin2022rap} leverage flat maxima to generate adversarial examples, PGN \cite{ge2023pgn} adopts a first-order procedure to approximate the Hessian/vector product, largely improving computational efficiency. They are not only effective on the surrogate model but also more resilient when attacking unseen target models. These methods mitigate overfitting, via making adversarial examples lie in regions where the loss function remains stable across different models (e.g., flat maxima regions), improving attack success rates in black-box settings. 
However, achieving flat maxima requires not only attention to ``flatness'' but also preservation of the ``maxima'' to ensure sufficient attack strength during transfer attack, which poses a significant challenge.

\subsection{Inner-Iteration Sampling} 
To enhance the generalization capability of adversarial examples, recent methods have introduced inner-iteration sampling \cite{wang2021vmifgsm, qin2022rap, ge2023pgn}. 
Specifically, an inner iterative process is inserted before updating the examples. 
This process involves neighborhood sampling to acquire domain information during inner-iterations which are used to adjust gradients \cite{wang2021vmifgsm} or neighborhood search is incorporated to locate local minima and enhance their loss values \cite{qin2022rap}.
Furthermore, works like \cite{zhu2023gra, ge2023pgn} enhance the flatness of the loss landscape by averaging gradients from inner-iterations to update the examples.

Additionally, neighborhood sampling can facilitate examples to escape from sharp local maxima regions, thereby avoiding overfitting to the surrogate model. 
However, directly averaging the gradients from inner-iterations will reduce the maximum value of the loss surface, since random sampling produces unstable gradient directions that fail to consistently direct to flat regions with higher local maxima. Thus, enhancing inner-iteration sampling efficiency to stabilize the final gradient has become a critical challenge.

\section{Methodology}
\label{method}

\subsection{Preliminaries}

%\subsubsection{Adversarial Attacks}
\textbf{Transferable Adversarial Attacks.}
For a given example $(x, y)$, a surrogate model $f_\theta$ and multiple target models $f_{\varphi_{k}}$ for $k \in \{1,2,...,K\}$, the attacker's goal is to find  $x^{adv} = x + \delta$ using $f_\theta$ to make $f_{\varphi_{k}}(x^{adv}) \neq y$, where $\|\delta\|_p < \epsilon$, $\epsilon$ denotes the maximum magnitude of the specified perturbation and $\|\cdot\|_p$ denotes the $\ell_p$-norm. For targeted attacks, we simply adapt the symbols and labels in the target function: $f_{\varphi_{k}}(x^{adv}) = y_t$, where $y_t$ denotes the target label. Meanwhile, to facilitate a clearer representation of the norm constraints against $x^{adv}$ as below, we define $\mathcal{B}_{\epsilon}(x)=\{x':\|x'-x\|_p \le \epsilon \}$ to denote the $\epsilon$-ball of an input image $x$. For the purpose of the above attack, we generally need to maximize the following objective function to generate the transferable adversarial examples:
    \begin{equation}
        \max _{x^{a d v} \in \mathcal{B}_{\epsilon}(x)} \mathcal{L}\left(x^{a d v}, y\right),
    \end{equation}
where $\mathcal{L}(\cdot)$ represents the loss function. 
In the above flow, we assume that $f_\theta$ is known as a white-box model and $f_{\varphi_{k}}$ is unknown as a black-box model. 
Similar to the process of training a neural network, 
the white-box model can be seen as the training data, with the adversarial examples serving as the optimization parameters, while the black-box model acts as the test data.

\textbf{Nesterov Accelerated Gradient (NAG).}
Due to the similarity between model training task and adversarial example generation, improving performance in black-box settings can be viewed as enhancing model generalization in model training tasks. Consequently, some commonly used optimization methods, such as Momentum \cite{rumelhart1986moment} and Nesterov Accelerated Gradient (NAG) \cite{nesterov1983nag}, can be introduced to improve the generalization of adversarial examples across black-box models through a more stable optimization process. 

NI-FGSM \cite{lin2019nifgsm_sim} was the first to incorporate the principles of NAG into adversarial attack tasks.
This enables the algorithm to acquire directional information in advance, reducing oscillations and achieving stable convergence.
Specifically, momentum is computed at each step as follows:
\begin{equation}
    \begin{aligned}
        v_{t} &=\gamma \cdot v_{t-1} + \frac{\nabla_{x} \mathcal{L}\left(\tilde{x}_t, y\right)}{\big\| \nabla_{x} \mathcal{L}\big(\tilde{x}_t, y\big) \big\|_{1}}, \\
        \tilde{x}_t &=\underbrace{x_{t-1}^{adv}+\alpha\cdot\gamma\cdot v_{t-1}}_\text{lookahead point},
    \end{aligned}
    \label{eq:ni}
\end{equation}
where $v_{t}$ represents the momentum term, 
$x_{0}^{adv}=x$, $\alpha$ stands for the step size, $\gamma$ denotes the momentum decay factor, and $\left\| \cdot \right\|_{1}$ denotes the $\ell_1$-norm of the given variable. 
Unlike traditional momentum-based methods, NI-FGSM utilizes the estimated position of examples rather than their current actual position, during gradient computation, i.e., introducing a \textbf{lookahead} mechanism into the algorithm.

\newcommand{\circled}[1]{%
  \protect\tikz[baseline=(char.base)]{
    \protect\node[shape=circle,draw,inner sep=0.5pt] (char) {\textbf{\scriptsize #1}};
  }%
}

\begin{figure}[!tb]
    \centering
    \includegraphics[width=0.45\textwidth ]{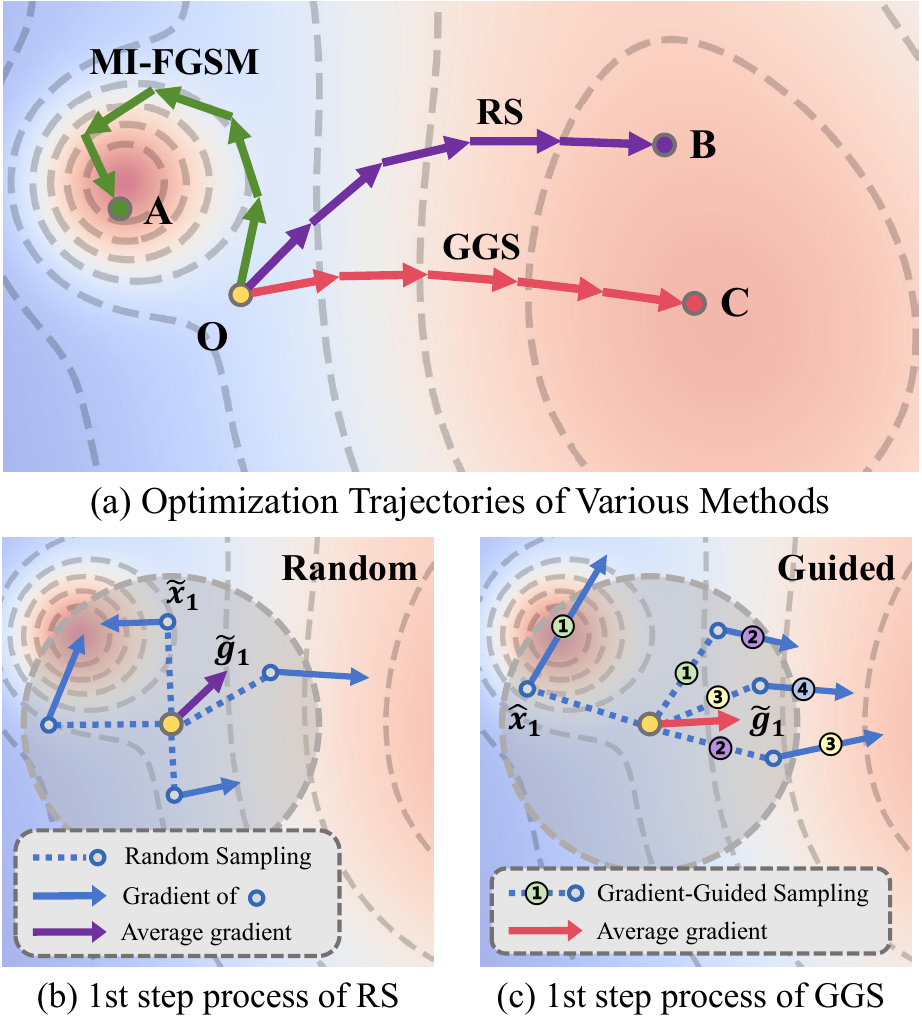}
    \caption{(a) %Algorithmic 
    Outer-Iteration processes of MI-FGSM (green), RS (purple) and our GGS (red). 
    (i) MI-FGSM prioritizes rapid ascent directions, thereby enhancing exploitation and facilitating access to sharp local maxima regions.
    (ii) RS introduces inner-iterations and enhances exploration capability through neighborhood search, thereby enabling access to flat local maxima regions.
    (iii) GGS incorporates gradient constraints into RS, simultaneously enhancing both exploration and exploitation capabilities, thereby enabling the faster convergence to the centers of flat local maxima regions.
    (b) RS is performed within a neighborhood of the current example, and uses the average gradient of all sampled points as the final gradient.
    (c) Building upon RS, GGS enables examples to have stable gradient directions (\circled{2} \circled{3} \circled{4}) toward the centers of flat local maxima regions, following an initial period of brief oscillation (\circled{1}). It uses the gradient direction from the previous inner-iteration as the guidance, while the randomness is maintained by setting the gradient magnitude with a random distribution. 
    }
    \label{fig:line_diagram}
\end{figure}

\subsection{Motivation}
\label{motivation}
Recent transfer-based adversarial attack methods have introduced inner-iteration random sampling, to enhance algorithm stability \cite{zhu2023gra, ge2023pgn} or  acquire neighborhood information \cite{wang2021vmifgsm, qin2022rap}. 
It can be noted that these methods have increased the flatness of the loss surface, as illustrated in Fig.\ref{fig:first_img}(c), with additional loss surface plots for other methods provided in the supplementary materials.
If we isolate the inner-iteration random sampling (RS) component, it can be expressed in the following form:
\begin{equation}
    \begin{aligned}
        v_{t} &= \gamma v_{t-1} + \frac{\sum_{i=1}^{N}\tilde{g}_{i}}{\left\|\sum_{i=1}^{N}\tilde{g}_{i}\right\|_{1}}, \\
        \tilde{g}_{i} &= \nabla_{x} \mathcal{L}\big(\tilde{x}_i, y\big), \:\,\text{with}\:\,
        \tilde{x}_{i} = \underbrace{x_{t-1}^{adv} + \tilde{p}}_\text{sampling point},
    \end{aligned}
    \label{eq:rs}
\end{equation}
where $\sum_{i=1}^{N}$ denotes the output of the inner-iteration process;
$N$ denotes the number of inner-iterations; $\gamma$ denotes the momentum decay factor;
$\tilde{g}_0, \tilde{p} \sim  \text{Uniform}(-\zeta, \zeta )$, are uniform random noises with the same size as $x$.

As shown in Fig. \ref{fig:first_img}, the inner-iteration RS possesses a strong capability to enhance the flatness of the loss surface compared to MI-FGSM. 
Compared to RS,
current methods such as PGN \cite{ge2023pgn} can further improve the flatness of the loss surface, however, they unwillingly reduce the local maxima of the loss surface, limiting its attack potential.
This reduction in local maxima can be attributed to the unstable gradients generated by the completely random sampling strategy, 
as illustrated in Fig. \ref{fig:line_diagram}(b). 
Specifically, the averaged gradient roughly points towards flat regions, while it struggles to consistently align with the stable direction to the center of flat maxima regions, thereby reducing updating efficiency.

To acquire stable gradient ascent direction and improve the efficiency of inner-iteration sampling, 
we define two essential characteristics that a well-balanced gradient direction should possess: 
(1) alignment with the stable gradient ascent direction for improving Exploitation capacity, and 
(2) pointing to loss regions with flatter surface for improving Exploration capacity.

\begin{figure}[!tb]
    \centering
    \includegraphics[width=0.48\textwidth ]{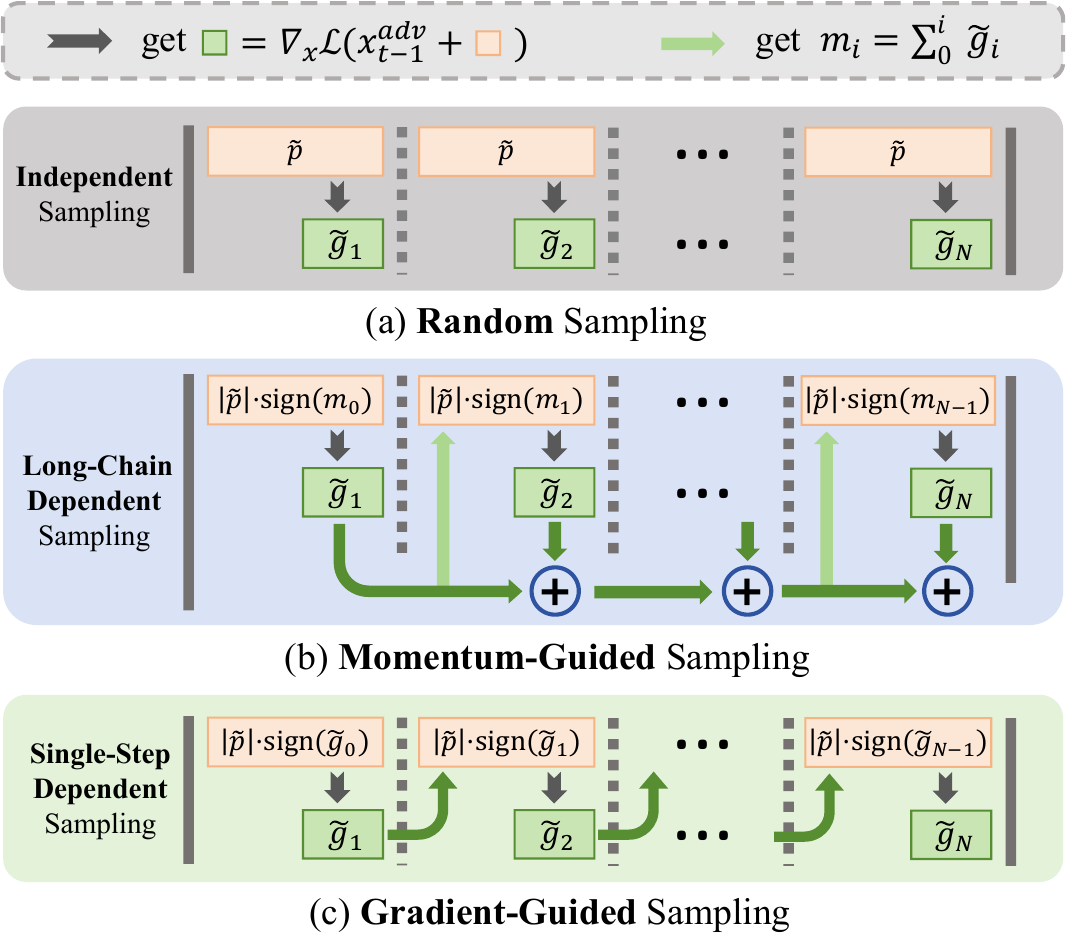} % 修改路径和尺寸
    \caption{
    The differences among three inner-iteration sampling guidance methods: 
    (a) Random sampling in Eq. \ref{eq:rs}, with each inner-iteration sampling being independent;
    (b) Momentum-guided sampling in Eq. \ref{eq:mrs}, where sampling direction depends on the cumulative average of all previous gradients, creating long-chain dependencies, while using random sampling for maintaining the randomness;
    (c) Gradient-guided sampling in Eq. \ref{eq:ggi}, where sampling direction relies solely on the gradient direction of the previous iteration, establishing single-step dependencies.
    }
    \label{fig:method}
\end{figure}

\subsection{Gradient-Guided Sampling}

\textbf{(1) Stable gradient ascent direction:} % Regarding the first point, 
A sampled gradient in a sufficiently small neighborhood of the current example is probably a gradient ascent direction, while how can we ensure that this gradient ascent direction is stable? 
The Nesterov Accelerated Gradient (NAG) technique can precisely address this issue. 
It first allows the example to ``lookahead" along the current momentum direction, and then acquires the gradient based on this new position, thereby identifying a more stable ascent direction \cite{nesterov1983nag, lin2019nifgsm_sim}. To leverage the superior stability of NAG based on random sampling, we perform the  magnitude sampling along the momentum direction with an uniform distribution, and named this approach as Momentum-Guided Sampling (MGS).
Compared to RS in Eq. \ref{eq:rs}, we only need to replace $\tilde{x}_i$ with $\bar{x}_i$ as follows:
\begin{equation}
    \bar{x}_{i}=\underbrace{x_{t-1}^{adv}+ \left|\tilde{p}\right| \cdot \text{sign}(m_{i-1})}_\text{lookahead sampling point},
    \label{eq:mrs}
\end{equation}
where \(\left|\tilde{p}\right|\) denotes the element-wise absolute value of \(\tilde{p}\), $\text{sign}(\cdot)$ denotes the sign function, which outputs either 1 or -1, and momentum decay is set to 1 for simplicity, so that $m_i=\sum_{k=1}^{i}\tilde{g}_{k}$.

\textbf{(2) Pointing to regions with better flatness:} %Regarding the second point, 
As shown in Fig. \ref{fig:method}(b), 
if we use momentum for `lookahead', 
the early unstable sampling in the inner-iterations will impose excessive constraints on subsequent sampling directions, leading to long-chain dependencies that severely impair RS's ability in exploring flat regions.
Thus, for our GGS, we substitute the momentum term in Eq. \ref{eq:mrs} with the gradient obtained from the previous iteration, where 
the lookahead mechanism is maintained with the gradient ascent direction.
Specifically, we only need to replace $\tilde{x}_i$ in Eq. \ref{eq:rs} with $\hat{x}_i$ as follows:
\begin{equation}
    \hat{x}_{i}=\underbrace{x_{t-1}^{adv}+ \left|\tilde{p}\right| \cdot \text{sign}(\tilde{g}_{i-1})}_\text{lookahead sampling point},
    \label{eq:ggi}
\end{equation}

Fig. \ref{fig:method} further sheds light on the differences among three inner-iteration sampling guidance methods.
As shown in Fig. \ref{fig:method}(c), GGS largely alleviates long-chain dependencies compared with Momentum-Guided Sampling in Fig. \ref{fig:method}(b), while ensuring the sampling direction to be align with the gradient ascent direction, compared with random sampling in Fig. \ref{fig:method}(a). 
In addition, as shown in Fig. \ref{fig:line_diagram}(c), GGS will achieve stable sampling directions (\circled{2} \circled{3} \circled{4}) after a brief period of oscillation (\circled{1}), when the initial sampling points are unstable.

For clarity, our attack process is shown in Algorithm \ref{algorithm:attack_new}.

\begin{algorithm}[htbp]
\caption{The attacking process of our GGS.}
    \begin{algorithmic}[1]
        \Require
            A clean image $x^{C \times H \times W}$ with ground-truth label $y$, and the loss function $\mathcal{L}(\cdot)$;
            the magnitude of perturbation $\epsilon$; 
            the sampling radius $\zeta$;
            the number of outer iterations, $T$; the decay factor $\gamma$; the number of inner iterations, $N$; 
            $\tilde{g}_{0}, \tilde{p} \in \mathbb{R}^{C \times H \times W}$
        \Ensure 
             An adversarial example $x^{adv}$;
        \State Randomly sample $\tilde{g}_{0} \sim  \text{Uniform}( - \zeta, \zeta )$;
        \State $v_{0}=0$, $x_0^{adv}=x$, $\alpha = \epsilon / T$; 
        \While {$t \gets 1$ to $T$ (outer-iteration)}
        \While {$i \gets 1$ to $N$ (inner-iteration)}
            \State Randomly sample noises $\tilde{p} \sim  \text{Uniform}( -\zeta, \zeta )$;
            \State Get the lookahead example $\hat{x}_i$ by Eq. \eqref{eq:ggi};
            \State Get the gradient $\tilde{g}_i=\nabla_{x} \mathcal{L}\big(\hat{x}_i, y\big)$
        \EndWhile
        \State Update $v_{t} = \gamma \cdot v_{t-1} + \frac{\sum_{i=1}^{N}\tilde{g}_{i}}{\left\|\sum_{i=1}^{N}\tilde{g}_{i}\right\|_{1}}$
        \State Update $x_{t}^{a d v}=\Pi_{\mathcal{B}_{\epsilon}(x)}\left[x_{t-1}^{a d v}+\alpha \cdot \operatorname{sign}\left(v_{t}\right)\right]$;
        \EndWhile
        \State $x^{a d v}=x_{T}^{a d v}$
    \end{algorithmic}
    \label{algorithm:attack_new}
\end{algorithm}

\section{Experiments}
\label{exp}

\begin{table*}
    \caption{The average untargeted and targeted ASR ($\%$) on the nine models (Res50 \cite{he2016resnet}, Dense121 \cite{huang2017densenet}, Inc-v3 \cite{szegedy2016incv3}, IncRes-v2 \cite{szegedy2017incresv2}, ViT-B \cite{dosovitskiy2020vit}, PiT-B \cite{heo2021pit}, and $\text{Inc-v3}_{ens3}$, $\text{Inc-v3}_{ens4}$, $\text{IncRes-v2}_{ens}$ \cite{tramer2017incens}), with adversarial examples generated on a single model (Res50, Inc-v3 or ViT-B). Each data pair ($u/w$) corresponds to the performances under (untargeted/targeted) attacks. The best and second best results are labeled in \textbf{bold} and \underline{underline}, respectively.}
    
    \label{result-table}
    \centering
    % \small
    % \tiny
    \footnotesize
    % \scriptsize
    
    % \begin{tabular}{l|c|cccccc|ccc|c}
    \begin{tabular}{@{\hspace{0.1em}}l@{\hspace{0.1em}}|
    @{\hspace{0.3em}}c@{\hspace{0.3em}}|
    @{\hspace{0.3em}}c@{\hspace{0.5em}}c@{\hspace{0.4em}}c@{\hspace{0.4em}}c@{\hspace{0.4em}}c@{\hspace{0.4em}}c
    @{\hspace{0.6em}}|@{\hspace{0.5em}}c@{\hspace{0.8em}}c@{\hspace{0.2em}}c@{\hspace{0.2em}}|
    @{\hspace{0.7em}}c@{\hspace{0.7em}}}
        \midrule \multirow{2}{*}{} 
        & Attack & \textbf{Res50} & Dense121 & \textbf{Inc-v3} & IncRes-v2 & \textbf{ViT-B} & PiT-B
        & $\text{Inc-v3}_{\text{ens3}}$ & $\text{Inc-v3}_{\text{ens4}}$ & $\text{IncRes-v2}_{\text{ens}}$ & Avg.\\
        
        \midrule \multirow{9}{*}{ Res50 } 
        & $\textbf{MI}_{\text{ CVPR'18}}$   & 99.8 / 98.1 & 54.9 / 0.2 & 44.2 / 0.0 & 28.0 / 0.0 & 11.8 / 0.0 & 22.4 / 0.0 & 22.8 / 0.0 & 25.3 / 0.0 & 19.6 / 0.0 & 36.53 / 10.92 \\
        & $\textbf{NI}_{\text{ ICLR'20}}$   & 100.0 / 95.1 & 63.1 / 0.2 & 47.2 / 0.0 & 32.3 / 0.0 & 12.9 / 0.0 & 23.7 / 0.0 & 23.0 / 0.0 & 27.5 / 0.0 & 19.0 / 0.0 & 38.74 / 10.59 \\
        & $\textbf{VMI}_{\text{ CVPR'21}}$  & 99.8 / 96.6 & 71.2 / 2.8 & 58.1 / 0.5 & 49.4 / 0.2 & 31.3 / 0.1 & 47.7 / 0.5 & 43.9 / 0.0 & 45.4 / 0.1 & 37.7 / 0.1 & 53.83 / 11.21 \\
        & $\textbf{RAP}_{\text{ NeurIPS'22}}$  & 99.6 / 26.4 & 86.8 / 0.5 & 68.4 / 0.0 & 52.6 / 0.0 & 22.2 / 0.0 & 40.5 / 0.1 & 36.2 / 0.0 & 36.7 / 0.0 & 25.5 / 0.0 & 52.06 / 3.00 \\
        & $\textbf{GRA}_{\text{ ICCV'23}}$  & 96.9 / 67.7 & 88.6 / 3.9 & 81.8 / 0.9 & 75.8 / 1.7 & 45.3 / 0.2 & 62.6 / 0.6 & 71.5 / 0.7 & 70.9 / 1.1 & 67.3 / 0.7 & 73.41 / 8.61 \\
        & $\textbf{PGN}_{\text{ NeurIPS'23}}$  & 98.6 / 49.5 & 91.3 / 4.7 & 85.0 / 1.4 & 78.5 / 1.7 & 49.7 / 0.4 & 67.8 / 1.0 & 74.9 / 0.6 & 72.9 / 1.1 & 70.1 / 1.2 & \underline{76.53} / 6.84 \\
        & $\textbf{ANDA}_{\text{ CVPR'24}}$ & 99.9 / 97.6 & 87.5 / 5.4 & 73.6 / 0.4 & 66.3 / 0.8 & 42.3 / 0.1 & 61.8 / 1.1 & 51.0 / 0.1 & 52.9 / 0.1 & 46.7 / 0.1 & 64.67 / \underline{11.74} \\
        & $\textbf{GI}_{\text{ ESWA'24}}$ & 100.0 / 98.5 & 68.7 / 0.4 & 55.6 / 0.1 & 37.6 / 0.1 & 15.3 / 0.0 & 29.0 / 0.0 & 28.8 / 0.0 & 30.4 / 0.1 & 22.4 / 0.0 & 43.09 / 11.02 \\
        & $\textbf{GGS}$ & 99.3 / 90.7 & 95.9 / 28.6 & 89.6 / 7.2 & 86.4 / 8.6 & 60.2 / 3.4 & 81.0 / 8.5 & 77.5 / 3.7 & 77.4 / 4.3 & 71.4 / 4.0 & \textbf{82.08} / \textbf{17.67} \\

        \midrule \multirow{9}{*}{Inc-v3}
        & MI   & 32.1 / 0.0 & 49.6 / 0.1 & 100.0 / 95.3 & 56.1 / 0.1 & 13.0 / 0.0 & 17.3 / 0.0 & 34.3 / 0.0 & 38.0 / 0.0 & 26.3 / 0.0 & 40.74 / \underline{10.61} \\
        & NI   & 41.4 / 0.0 & 60.5 / 0.1 & 100.0 / 75.9 & 66.9 / 0.1 & 13.4 / 0.0 & 19.2 / 0.0 & 36.8 / 0.0 & 40.6 / 0.1 & 29.5 / 0.0 & 45.37 / 8.47 \\
        & VMI  & 44.9 / 0.1 & 65.6 / 0.1 & 99.9 / 76.6 & 72.7 / 0.4 & 18.3 / 0.0 & 25.3 / 0.0 & 48.9 / 0.1 & 52.5 / 0.0 & 40.9 / 0.0 & 52.11 / 8.59 \\
        & RAP  & 58.0 / 0.0 & 78.4 / 0.2 & 100.0 / 16.4 & 82.1 / 0.1 & 15.4 / 0.0 & 23.8 / 0.0 & 44.1 / 0.1 & 47.0 / 0.0 & 33.9 / 0.0 & 53.63 / 1.87 \\
        & GRA  & 53.3 / 0.1 & 75.4 / 0.4 & 99.8 / 67.5 & 85.4 / 1.8 & 21.4 / 0.0 & 30.5 / 0.1 & 64.9 / 0.1 & 65.9 / 0.3 & 56.1 / 0.1 & 61.41 / 7.82 \\
        & PGN  & 56.0 / 0.0 & 79.4 / 0.2 & 100.0 / 50.3 & 87.9 / 0.9 & 23.2 / 0.0 & 33.4 / 0.0 & 65.9 / 0.0 & 67.0 / 0.1 & 55.6 / 0.0 & \underline{63.16} / 5.72 \\
        & ANDA & 51.3 / 0.3 & 76.3 / 0.4 & 100.0 / 87.4 & 82.5 / 0.6 & 20.9 / 0.0 & 31.8 / 0.1 & 58.2 / 0.2 & 61.2 / 0.2 & 49.3 / 0.0 & 59.06 / \underline{9.91} \\
        & GI   & 39.9 / 0.0 & 59.3 / 0.3 & 100.0 / 97.1 & 62.6 / 0.5 & 14.6 / 0.0 & 19.9 / 0.0 & 39.2 / 0.0 & 40.9 / 0.0 & 31.9 / 0.0 & 45.37 / \textbf{10.88} \\
        & GGS  & 69.1 / 1.1 & 86.9 / 1.9 & 100.0 / 76.0 & 95.8 / 6.5 & 27.9 / 0.0 & 39.7 / 0.1 & 70.0 / 0.7 & 73.9 / 1.4 & 60.1 / 0.5 & \textbf{69.27} / 9.80 \\

        \midrule \multirow{9}{*}{ ViT-B } 
        & MI   & 41.7 / 0.0 & 55.6 / 0.0 & 58.1 / 0.0 & 39.4 / 0.0 & 100.0 / 99.9 & 50.1 / 0.0 & 34.2 / 0.0 & 36.3 / 0.0 & 29.9 / 0.0 & 49.48 / 11.10 \\
        & NI   & 42.4 / 0.0 & 58.3 / 0.0 & 59.5 / 0.0 & 44.7 / 0.1 & 100.0 / 99.0 & 50.0 / 0.2 & 36.2 / 0.0 & 37.6 / 0.0 & 31.0 / 0.0 & 51.08 / 11.03 \\
        & VMI  & 54.8 / 0.6 & 65.9 / 0.4 & 63.1 / 0.1 & 50.9 / 0.3 & 100.0 / 99.0 & 67.6 / 1.5 & 45.3 / 0.1 & 44.7 / 0.0 & 38.7 / 0.2 & 59.00 / 11.36 \\
        & RAP  & 62.1 / 0.1 & 79.6 / 0.2 & 73.7 / 0.0 & 62.3 / 0.0 & 99.9 / 57.0 & 66.2 / 0.1 & 48.5 / 0.0 & 51.2 / 0.0 & 42.6 / 0.0 & 65.12 / 6.38 \\
        & GRA  & 64.9 / 1.0 & 77.6 / 1.1 & 74.7 / 0.8 & 68.8 / 0.8 & 99.3 / 91.7 & 80.9 / 4.7 & 61.5 / 0.2 & 64.0 / 0.1 & 57.7 / 0.2 & 72.16 / 11.18 \\
        & PGN  & 69.3 / 0.4 & 81.1 / 0.7 & 78.6 / 0.4 & 70.2 / 0.5 & 99.1 / 82.2 & 84.8 / 4.4 & 65.6 / 0.1 & 67.5 / 0.3 & 61.4 / 0.2 & \underline{75.29} / 9.91 \\
        & ANDA & 66.2 / 0.3 & 79.0 / 0.5 & 75.5 / 0.1 & 65.7 / 0.3 & 100.0 / 99.9 & 79.0 / 1.4 & 55.3 / 0.1 & 56.7 / 0.1 & 49.9 / 0.2 & 69.70 / \underline{11.44} \\
        & GI   & 55.8 / 0.1 & 69.6 / 0.2 & 67.1 / 0.0 & 52.5 / 0.1 & 100.0 / 100.0 & 61.0 / 0.3 & 45.4 / 0.0 & 46.3 / 0.0 & 38.0 / 0.0 & 59.52 / 11.19 \\
        & GGS  & 80.8 / 6.0 & 89.9 / 6.5 & 87.4 / 4.6 & 82.3 / 4.6 & 100.0 / 99.2 & 92.7 / 27.3 & 74.0 / 2.1 & 74.5 / 2.3 & 68.4 / 2.4 & \textbf{83.33} / \textbf{17.22} \\

        \midrule
    \end{tabular}
\end{table*}

% ============================== Experimental Settings ==============================

\subsection{Experimental Settings}
\label{exp:set}

% 1. Dataset
\textbf{Dataset:} Our experiments are conducted on an ImageNet-compatible dataset that is widely used in adversarial attack research. The dataset includes 1,000 images, each with a size of 299 × 299 × 3.
% 2. Models

\textbf{Models:} We select six normally pre-trained models, i.e., ResNet50 (Res50) \cite{he2016resnet}, DenseNet121 (Dense121) \cite{huang2017densenet}, Inception-v3 (Inc-v3) \cite{szegedy2016incv3}, InceptionResNet-v2 (IncRes-v2) \cite{szegedy2017incresv2}, VisionTransformer-Base (ViT-B) \cite{dosovitskiy2020vit}, and Pooling-based VisionTransformer-Base (PiT-B) \cite{heo2021pit}, along with three models pre-trained on adversarial examples $\text{Inception-v3}_{ens3}$ ($\text{Inc-v3}_{ens3}$), $\text{Inception-v3}_{ens4}$ ($\text{Inc-v3}_{ens4}$) and $\text{InceptionResNet-v2}_{ens}$ ($\text{IncRes-v2}_{ens}$) \cite{tramer2017incens}, to evaluate the performance of different methods.
% 3. Baselines

\textbf{Baselines:} To provide a comprehensive evaluation of our method, 
we choose eight popular gradient-based adversarial attack methods as baselines: MI-FGSM \cite{dong2018mifgsm}, NI-FGSM \cite{lin2019nifgsm_sim}, VMI-FGSM \cite{wang2021vmifgsm}, RAP \cite{qin2022rap}, GRA \cite{zhu2023gra}, PGN \cite{ge2023pgn}, ANDA \cite{fang2024anda} and GI \cite{wang2024gifgsm}. Additionally, we integrate our method with five input transformation-based methods, i.e., DIM \cite{xie2019dim}, TIM \cite{dong2019tim}, SIM \cite{lin2019nifgsm_sim}, Admix \cite{wang2021admix}, and SSM \cite{long2022ssm} to verify the generalizability of our approach.
% 4. Hyperparameters

\textbf{Hyperparameters:} For our GGS, we set the maximum perturbation $\epsilon = 16/255$, the number of outer-iterations $T = 10$, the step size $\alpha = \epsilon/T$, the number of inner-iterations $N = 20$, and the upper bound factor of sample range $\zeta = 2.0 \times \epsilon$. More hyperparameter explanation for other methods is provided in the Supplementary Material.

% ============================== Attack on Single and Ensemble Models ==============================
\subsection{Attack on Different Models}

\subsubsection{Classification model}

\label{exp:attack}

As shown in Table \ref{result-table}, our method consistently outperforms state-of-the-art (SOTA) approaches. Compared to the best-performing competitor, it improves the average attack success rate (ASR) by over 5\% in both untargeted and targeted attacks across most architectures, except for Inception-v3. Furthermore, our GGS significantly enhances attack performance on the surrogate model, achieving 4\%–25\% higher ASR in targeted attacks than other sampling methods like PGN \cite{ge2023pgn} and GRA \cite{zhu2023gra}. This demonstrates GGS's ability to boost attack potency while maintaining strong cross-model generalization. Additional details on ensemble setting can be found in the supplementary materials.

\begin{figure*}[!tb]
    % \vspace{-1em}
    \centering
    \includegraphics[width=1\textwidth ]{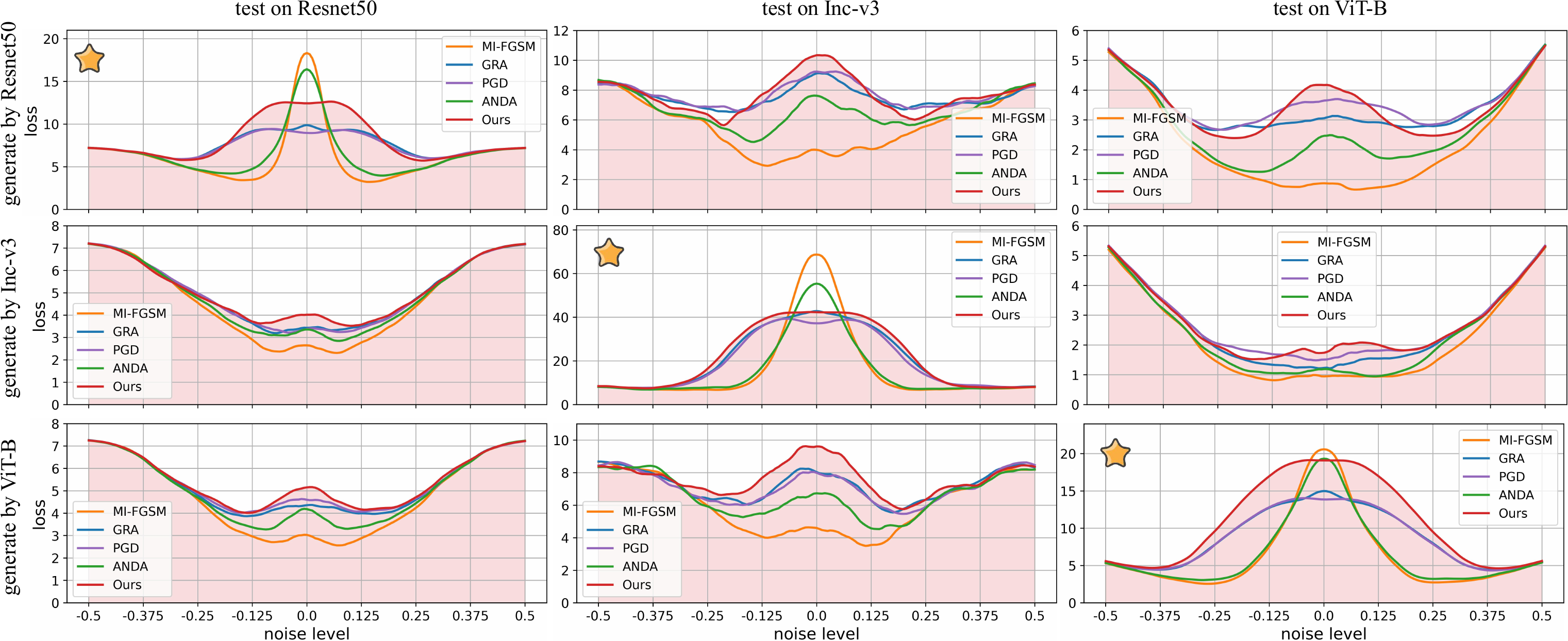} 
    \caption{The loss surfaces of adversarial examples generated by different methods (MI-FGSM \cite{dong2018mifgsm}, GRA \cite{zhu2023gra}, PGN \cite{ge2023pgn}, ANDA \cite{fang2024anda}), based on three different model architectures (ResNet50 \cite{he2016resnet}, Inception-v3 \cite{szegedy2016incv3}, and ViT-B \cite{dosovitskiy2020vit}) along random directions with varying strengths like \cite{ge2023pgn}.
    % , reveals critical insights. 
    In the images marked with a star in the upper left corner, adversarial examples are generated and tested on the same model, indicating white-box testing. In contrast, unmarked images represent adversarial examples generated on one model and tested on another, indicating black-box testing. We have also highlighted the regions covered by our loss surface using a red background for easier visualization. 
    }
    % \vspace{-1em}
    \label{fig:loss_surf_all}
\end{figure*}

% ============================== Attack MLLMS ==============================

\subsubsection{Multimodal Large Language Models}

The widespread adoption of Multimodal Large Language Models (MLLMs) has raised significant security concerns. We evaluated our method on five prominent MLLMs using adversarial examples generated under an ensemble setting, as shown in Table \ref{tab:llm}. Our GGS method achieved superior attack performance, reducing the average CSR by over 9\% compared to the strongest baseline, which demonstrates its excellent cross-model applicability.

\begin{table}[htbp]
\small 
\centering
    \caption{Classification success rate (CSR, \%) of adversarial examples generated by different methods in Multimodal Large Language Models (MLLMs) 
    from three major providers (OpenAI, Google, Anthropic) 
    under ensemble settings (Res50 \cite{he2016resnet}, Inc-v3 \cite{szegedy2016incv3}, ViT-B \cite{dosovitskiy2020vit}). Row ``clean" indicates the CSR on the clean samples. 
    Avg. represents the average CSR across the five MLLMs.
    }
    \vspace{-0.8em}
    \label{tab:llm}
    \setlength{\tabcolsep}{4pt}
    \begin{tabular}{c|cc|cc|c|c}
        \toprule    
        \multirow{2}{*}{Method} & \multicolumn{2}{c|}{GPT} & \multicolumn{2}{c|}{Gemini} & \multicolumn{1}{c|}{Claude} & \multirow{2}{*}{Avg. $\downarrow$} \\
             & 4o   & mini   & pro    & flash  & sonnet  &  \\
        \midrule
        Clean& 77.1 & 84.0 & 85.6 & 80.9 & 68.3 & 79.18\\
        \midrule
        $\textbf{MI}_{\text{ CVPR'18 \cite{dong2018mifgsm}}}$   & 77.0 & 77.3 & 81.8 & 76.5 & 61.0 & 74.72\\
        $\textbf{NI}_{\text{ ICLR'20 \cite{lin2019nifgsm_sim}}}$& 75.0 & 73.9 & 83.1 & 76.8 & 62.4 & 74.24\\
        $\textbf{VMI}_{\text{ CVPR'21 \cite{wang2021vmifgsm}}}$ & 68.8 & 66.6 & 78.4 & 73.2 & 63.1 & 70.02\\
        $\textbf{RAP}_{\text{ NeurIPS'22 \cite{qin2022rap}}}$   & 64.3 & 62.0 & 75.1 & 66.1 & 60.2 & 65.54\\
        
        $\textbf{GRA}_{\text{ ICCV'23 \cite{zhu2023gra}}}$    & 57.1 & 52.2 & 70.9 & 65.6 & 46.7 & 58.50\\
        $\textbf{PGN}_{\text{ NeurIPS'23 \cite{ge2023pgn}}}$  & \underline{56.2} & \underline{49.0} & \underline{69.6} & \underline{64.1} & \underline{45.1} & \underline{56.80}\\
        $\textbf{ANDA}_{\text{ CVPR'24 \cite{fang2024anda}}}$ & 59.9 & 56.8 & 74.9 & 69.3 & 50.5 & 62.28\\
        $\textbf{GI}_{\text{ ESWA'24 \cite{wang2024gifgsm}}}$ & 71.5 & 72.8 & 79.4 & 73.2 & 63.8 & 72.14\\
        
        \textbf{GGS}  & \textbf{43.1} & \textbf{37.8} & \textbf{61.1} & \textbf{55.7} & \textbf{40.0} & \textbf{47.54}\\
        \bottomrule
    \end{tabular}
\end{table}

% % ============================== 2D Loss Surfaces Visualization ==============================
% \section{2D Loss Surfaces Visualization}

\subsubsection{Loss Surface Visualization against Noise Levels}
To shed light on the capacity of our GGS in locating flat maxima regions for adversarial examples,
we compared the loss surfaces of adversarial examples generated by different attack methods in Fig. \ref{fig:loss_surf_all}. Each curve represents the average loss of 32 randomly selected adversarial examples, while the center of each image (noise level = 0) indicates the average loss of clean adversarial examples. In white-box settings (marked with a star), GGS consistently finds flatter maxima compared to baselines.
Consequently, in black-box testing, GGS achieves the highest loss values (images without star), indicating its superior transfer attack capability. 
Additionally, the GGS loss surface (highlighted in red) almost entirely encompasses those of other methods, confirming its effectiveness in balancing strong attack potency with cross-model generalization.

% ============================== Attack with Other Methods ==============================

\begin{figure*}[!t]
    \centering
    \includegraphics[width=1\textwidth ]{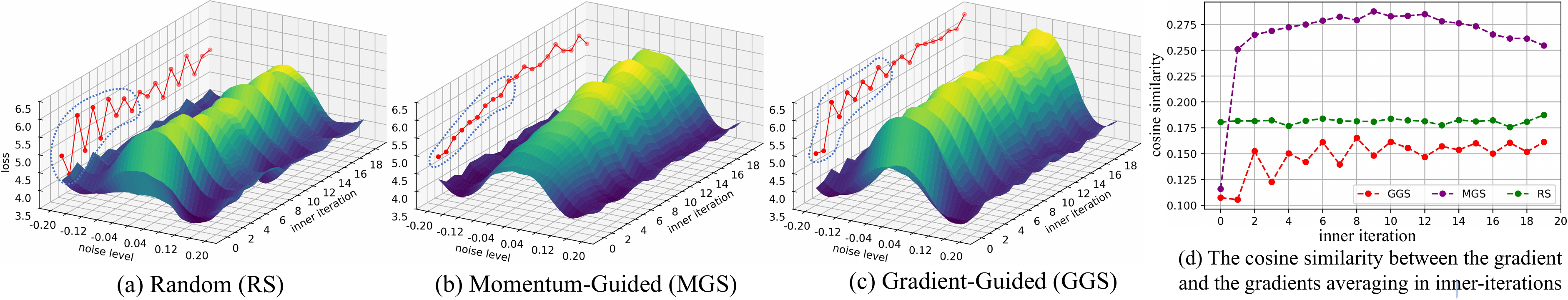} % 修改路径和尺寸
    \caption{Loss surfaces of adversarial examples generated by different sampling strategies, Random Sampling (RS), MGS and our GGS, with increasing inner-iteration on Resnet50. The red line on the left side of part (a)$\sim$(c) represents the maximum value of the loss surface in each iteration, and the blue dashed line highlight these values in early-stage inner iteration. For (d), it represents the cosine similarity between the gradient $\tilde{g}_{i}$ generated in each inner-iteration and the average of gradients $\sum_{i=1}^N\tilde{g}_{i}/N$.}
    \label{fig:ablation_momentum}
\end{figure*}

\begin{table}[htbp]
    \small
    \centering
    \caption{The average untargeted and targeted ASR ($\%$) on all nine models with the adversarial examples generated on ResNet50, when combined our GGS with input transformation-based methods (DIM \cite{xie2019dim}, TIM \cite{dong2019tim}, SIM \cite{lin2019nifgsm_sim}, Admix \cite{wang2021admix}, SSM \cite{long2022ssm}) and inner-iteration RS-based methods (GRA \cite{zhu2023gra}, PGN \cite{ge2023pgn}).}
    \label{tab:conbined}
    \setlength{\tabcolsep}{2.6pt}
    \begin{tabular}{c|c|cc}
        \toprule
        Class & Method & Untargeted Avg. & Targeted Avg. \\
        \midrule
        \multirow{2}{*}{\makecell{Random \\ Sampling (RS)}}
        & GRA / +ours     & 73.41 / \textbf{78.69} & 8.61 / \textbf{13.64} \\
        & PGN / +ours     & 76.53 / \textbf{83.23} & 6.84 / \textbf{12.79} \\
        \midrule
        \multirow{5}{*}{\makecell{Input \\ Transformation}}
        & DIM / +ours     & 51.30 / \textbf{90.13} & 8.63  / \textbf{19.37} \\
        & TIM / +ours     & 40.82 / \textbf{84.94} & 11.03 / \textbf{18.48} \\
        & SIM / +ours     & 46.80 / \textbf{90.12} & 11.14 / \textbf{28.44} \\
        & Admix / +ours   & 54.78 / \textbf{85.50} & 11.28 / \textbf{30.50} \\
        & SSM / +ours     & 74.22 / \textbf{80.74} & 8.40 / \textbf{12.54} \\
        \bottomrule
    \end{tabular}
\end{table}

% \vspace{-1em}
\subsection{Compatibility with Other Methods}
\label{exp:with}

\textbf{Attack with Inner-Iteration Sampling Methods:}
% Our Gradient-Guided Sampling (GGS) enhances the stability and transferability of existing inner-iteration random sampling methods. 
Our Gradient-Guided Sampling (GGS) is compatible with existing inner-iteration RS-based approaches.
When integrated with SOTA approaches like GRA \cite{zhu2023gra} and PGN \cite{ge2023pgn}, GGS largely improves attack success rates (ASR). As shown in Table \ref{tab:conbined}, untargeted ASR improved by 5.2\% (GRA) and 6.7\% (PGN), while targeted ASR increased by 5\% (GRA) and 6\% (PGN), demonstrating GGS’s compatibility.

\textbf{Attack with Input Transformations Methods:}
To study the compatibility of our GGS with existing input transformation techniques, 
%
%To evaluate GGS’s compatibility, 
we integrated it with five methods of this type: DIM \cite{xie2019dim}, TIM \cite{dong2019tim}, SIM \cite{lin2019nifgsm_sim}, Admix \cite{wang2021admix}, and SSM \cite{long2022ssm}. Adversarial examples generated on ResNet50 were tested on nine diverse models.
Table \ref{tab:conbined} shows GGS significantly improved adversarial transferability, increasing untargeted ASR by 6\%–43\% and targeted ASR by 4\%–19\%. For instance, 
SIM’s untargeted ASR improved by 43\% when combined with GGS.
More details are in the supplementary material.

% ============================== Component Ablation ==============================

\subsection{Ablation Study of Proposed Components}
\label{exp:ablation}
% \vspace{-1em}

\begin{table}[htbp]
    \small
    \centering
    \caption{
    The average untargeted ASR on ResNet50 and eight target models for Random Sampling (RS), Momentum-Guided Sampling (MGS), and our Gradient-Guided Sampling (GGS).
    }
    \label{tab:ablation}
    \setlength{\tabcolsep}{5pt}
    \begin{tabular}{l|ccc|cc}
        \toprule
        & \multicolumn{3}{c|}{Sampling Guided by} & \multirow{2}{*}{\makecell{Res50}} & \multirow{2}{*}{\makecell{Others}} \\
        % \cline{2-6}
        & rand & momentum & gradient & & \\
        \midrule
        (a) RS  & {\checkmark} & - & - & 97.3  & \underline{63.74} \\
        (b) MGS & - & {\checkmark} & - & \underline{97.5} & 58.79 \\
        (c) GGS & - & - & {\checkmark} & \textbf{99.3}  & \textbf{79.93} \\
        \bottomrule
    \end{tabular}
\end{table}

Ablation study compares RS, MGS, and GGS via loss surface analysis (Fig. \ref{fig:ablation_momentum}) and attack success rates (Table \ref{tab:ablation}). 

As shown in Fig. \ref{fig:ablation_momentum}(a), Random Sampling (RS) yields a relatively flat loss surface due to its exploration capability but it suffers from significant oscillation in loss values. Introducing Momentum-Guided Sampling (MGS) (Fig. \ref{fig:ablation_momentum}(b)) stabilizes the loss surface over iterations, but its long-chain dependency causes high similarity between later gradients and the final average gradient (Fig. \ref{fig:ablation_momentum}(d)), severely limiting exploration. Our GGS resolves MGS’s long-chain dependency and mitigates early-stage noisy gradients (Fig. \ref{fig:ablation_momentum}(b)(c), blue dashed lines). The low gradient similarity in Fig. \ref{fig:ablation_momentum}(d) reflects the enhanced exploration, while the flat yet higher-loss surface in Fig. \ref{fig:ablation_momentum}(c) demonstrates the exceptional sampling efficiency of our GGS.

Quantitatively, as shown in Table \ref{tab:ablation}, MGS marginally improves ASR of the surrogate model by 0.2\%, but reducing transfer attack capability by 5\%. In contrast, GGS increases both white-box and transfer ASR by 2\% and 16\%, respectively, validating its superior sampling efficiency.

\section{Conclusion and Perspectives}
% \section{Conclusion}
\label{sec:conclusion}

This paper introduces Gradient-Guided Sampling (GGS), an easy-to-implement inner-iteration sampling strategy, for transferable adversarial attacks that effectively resolves the dilemma between Exploitation (attack potency) and Exploration (cross-model generalization). By guiding the sampling direction with the gradient from the previous inner iteration, GGS enables stable ascent toward flat loss regions with higher local maxima. Our GGS has the merit of enhancing exploitation capability without compromising exploration capability, effectively balancing these two aspects and addressing the inherent limitations of random sampling (RS) methods, as verified by extensive experiments and visualizations as well as the comparison with related state-of-the-art methods. GGS is also revealed to be highly compatible with existing methods.

While our GGS is highly compatible with gradient-averaging methods (e.g., GRA \cite{zhu2023gra}, PGN \cite{ge2023pgn}), we aim to refine it 
%into a foundational method supporting 
to also support non-gradient-averaging techniques (e.g., VMI-FGSM \cite{wang2021vmifgsm}, RAP \cite{qin2022rap}).

\section*{Acknowledgements}
The work was supported by the National Natural Science Foundation of China under grants no. 62276170, 82261138629, 62306061, 
the Science and Technology Project of Guangdong Province under grants no. 2023A1515010688, 
the Science and Technology Innovation Commission of Shenzhen under grant no. JCYJ20220531101412030, 
Open Research Fund from Guangdong Laboratory of Artificial Intelligence and Digital Economy (SZ) under Grant No. GML-KF-24-11, 
and Guangdong Provincial Key Laboratory under grant no. 2023B1212060076.

{
    \small
    \bibliographystyle{ieeenat_fullname}
    \bibliography{main}
}

\clearpage
\setcounter{page}{1}

\maketitlesupplementary

% ============================== Robustness of Defense Methods ==============================

\section{Robustness against Defense Methods}
\label{exp:defense}

To further study the robustness of our method, we selected seven defense strategies for testing, i.e., RP \cite{xie2017rp}, Bit-red \cite{xu2017bitred}, JPEG \cite{dziugaite2016jpeg}, FD \cite{liu2019fd}, NRP \cite{naseer2020nrp}, DP (DiffPure) \cite{nie2022diffpure} and MD (MimicDiffusion) \cite{song2024mimicdiffusion}.
Table \ref{tab:defense} presents the ASR of various methods under these defenses.
Table \ref{tab:defense} indicates that GGS demonstrates superior performance in overcoming defenses except for DP and NRP, increasing the average ASR by $1.7\%$ compared to state-of-the-art methods, validating its robustness and suggesting its greater applicability and effectiveness in a wide range of scenarios.

Moreover, when comparing the inner-iteration methods using gradient averaging (GRA \cite{zhu2023gra}, PGN \cite{ge2023pgn}, GGS) with other methods, the minimum difference in average attack success rate is 12\% (ANDA:54.2\% to GRA: 66.74\%), demonstrating the outstanding effectiveness of inner-iteration sampling methods in overcoming defense mechanisms.

\begin{table}[htbp]
    % \scriptsize
    % \footnotesize
    \small
    % \footnotesize
    \centering
    \caption{The average untargeted ASR ($\%$) on nine models under seven defense strategies (RP \cite{xie2017rp}, Bit-red \cite{xu2017bitred}, JPEG \cite{dziugaite2016jpeg}, FD \cite{liu2019fd}, NRP \cite{naseer2020nrp}, DP \cite{nie2022diffpure} and MD \cite{song2024mimicdiffusion}) of different methods (MI\cite{dong2018mifgsm}, NI\cite{lin2019nifgsm_sim}, VMI\cite{wang2021vmifgsm}, RAP \cite{qin2022rap}, GRA\cite{zhu2023gra}, PGN\cite{ge2023pgn}, ANDA\cite{fang2024anda}, GI \cite{wang2024gifgsm}, GGS), with the adversarial examples generated on ResNet50.}
    \label{tab:defense}
    \setlength{\tabcolsep}{3.2pt}
    \begin{tabular}{c|ccccc|cc|c}
        \toprule % DiffPure, MimicDiffusion
        Method & RP & Bit-red & JPEG & FD & NRP & DP & MD & Avg.\\ % Mimic-Diffusion 
        \midrule
        MI   & 37.8 & 36.2 & 33.5 & 41.0 & 30.8 & 19.2 & 37.5 & 33.71 \\
        NI   & 39.9 & 38.0 & 34.3 & 42.6 & 31.3 & 19.6 & 38.1 & 34.83 \\
        VMI  & 54.0 & 52.9 & 49.7 & 53.5 & 40.8 & 28.0 & 50.1 & 47.00 \\
        RAP  & 53.7 & 51.8 & 49.0 & 36.6 & 43.0 & 26.1 & 50.5 & 44.39 \\
        GRA  & 74.1 & 72.9 & 72.8 & 72.6 & 59.9 & \underline{46.2} & 68.7 & 66.74 \\
        PGN  & \underline{77.0} & \underline{76.3} & \underline{76.0} & \underline{75.2} & \textbf{61.3} & \textbf{47.0} & \underline{72.4} & \underline{69.31} \\
        ANDA & 66.0 & 62.8 & 57.9 & 63.2 & 41.2 & 31.1 & 57.2 & 54.20 \\
        GI   & 44.7 & 43.0 & 40.6 & 34.0 & 47.53 & 21.8 & 42.7 & 39.19 \\
        GGS  & \textbf{82.2} & \textbf{81.3} & \textbf{79.0} & \textbf{78.3} & \underline{60.6}  & 42.1 & \textbf{73.9} & \textbf{71.06}\\
        \bottomrule
    \end{tabular}
\end{table}

% ============================== Attack on Cloud Models ==============================

\section{Attack on Cloud Models}

\begin{table}[htbp]
\small 
\centering
    \caption{Classification success rate (\%) of adversarial examples generated by all nine methods (MI-FGSM \cite{dong2018mifgsm}, NI-FGSM \cite{lin2019nifgsm_sim}, VMI-FGSM \cite{wang2021vmifgsm}, RAP \cite{qin2022rap}, GRA \cite{zhu2023gra}, PGN \cite{ge2023pgn}, ANDA \cite{fang2024anda}, GI-FGSM \cite{wang2024gifgsm}, GGS) in the general label classification interfaces provided by four prominent cloud service providers (Alibaba, Tencent, Baidu, HUAWEI). The row ``clean'' indicates the classification accuracy corresponding to the original, unperturbed samples.
    }
    \label{tab:cloud}
    \setlength{\tabcolsep}{3.2pt}
    \begin{tabular}{c|cccc|c}
        \toprule
        Method & Alibaba & Tencent  & Baidu & HUAWEI & Avg. $\downarrow$ \\
             % & 4o   & mini   & pro    & flash  & sonnet  \\
        \midrule
        clean                                                   & 81.3 & 47.0 & 48.5 & 64.3 & 60.28 \\
        \midrule
        $\textbf{MI}_{\text{ CVPR'18 \cite{dong2018mifgsm}}}$   & 44.8 & 24.7 & 29.7 & \underline{25.0} & 31.05 \\
        $\textbf{NI}_{\text{ ICLR'20 \cite{lin2019nifgsm_sim}}}$& 40.3 & 26.4 & 26.4 & 49.8 & 35.73 \\
        $\textbf{VMI}_{\text{ CVPR'21 \cite{wang2021vmifgsm}}}$ & 35.4 & 22.8 & 23.6 & 47.0 & 32.20 \\
        $\textbf{RAP}_{\text{ NeurIPS'22 \cite{qin2022rap}}}$   & 22.6 & 20.6 & 18.8 & 47.2 & 27.30 \\
        $\textbf{GRA}_{\text{ ICCV'23 \cite{zhu2023gra}}}$      & 25.3 & 17.5 & 14.6 & 34.2 & 22.90 \\
        $\textbf{PGN}_{\text{ NeurIPS'23 \cite{ge2023pgn}}}$    & \underline{22.2} & \underline{17.5} & \underline{14.5} & 33.8 & \underline{22.00} \\
        $\textbf{ANDA}_{\text{ CVPR'24 \cite{fang2024anda}}}$   & 24.0 & 17.9 & 17.3 & 38.1 & 24.33 \\
        $\textbf{GI}_{\text{ ESWA'24 \cite{wang2024gifgsm}}}$   & 37.8 & 27.2 & 28.6 & 48.9 & 35.63 \\
        \textbf{GGS}                                            & \textbf{15.7} & \textbf{13.9} & \textbf{10.9} & \textbf{21.7} & \textbf{15.55} \\
        \bottomrule
    \end{tabular}
\end{table}

To further evaluate the capabilities of different methods in real-world scenarios, we selected four prominent cloud service providers: Alibaba, Tencent, Baidu, and HUAWEI, and conducted attack tests on their general image labeling services. As shown in Table \ref{tab:cloud}, our GGS achieved the best attack performance across all the four providers, with the final average result being 6\% lower than current best result, demonstrating the effectiveness of our GGS in practical applications.

% ============================== Visualization of Loss Surfaces against Inner Iteration.==============================

\begin{figure*}[!tb]
    \centering
    \includegraphics[width=1\textwidth ]{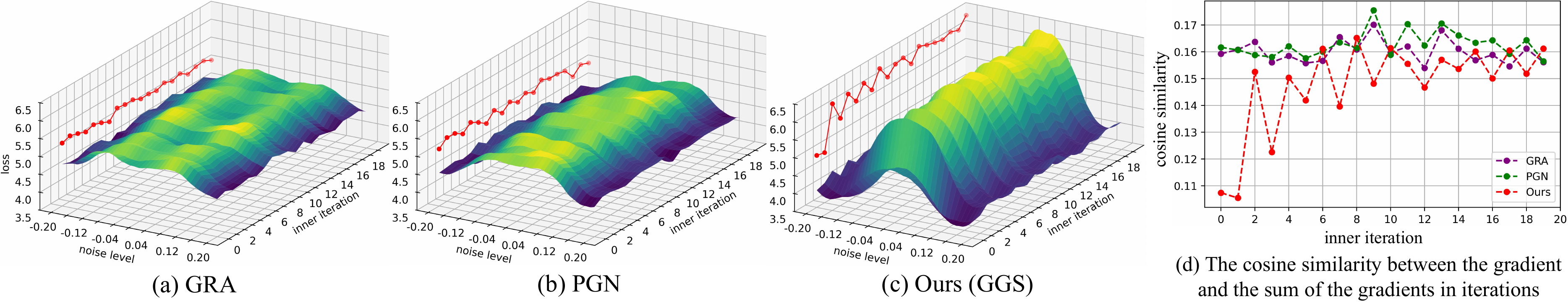} % 修改路径和尺寸
    \caption{Loss surfaces of adversarial examples generated by GRA\cite{zhu2023gra}, PGN\cite{ge2023pgn}, and our GGS, with increasing inner-iterations on Resnet50. The red line on the right side of part (a)$\sim$(c) represents the maximum value of the loss surface in each iteration. For (d), it represents the cosine similarity between the gradient $\tilde{g}_i$ generated in each inner-iteration and the average of gradients $\sum_{i=1}^N\tilde{g}_i/N$.}
    \label{fig:in3d_loss_surf}
\end{figure*}

% Visualization of 
\section{Loss Surfaces against Inner-Iteration}
\label{exp:visual}

To further compare the differences in the inner-iteration processes between our method and other inner-iteration sampling methods, we visualized the loss surface  against different inner-iteration steps of PGN \cite{ge2023pgn}, GRA \cite{zhu2023gra}, and GGS, as well as the similarity between the gradients during the inner-iterations and the final gradient averaging, as shown in Fig. \ref{fig:in3d_loss_surf}.

In Fig. \ref{fig:in3d_loss_surf}(d), it shows that the cosine similarity between the generated gradients and the final averaged gradient specific to our GGS gradually increases and stabilizes, as the number of inner-iteration increases, 
indicating that our method tends to favor the more stable gradient results from the later-stages of inner-interations, 
whereas PGN and GRA do not exhibit significant differential treatment toward different gradients.

From Fig. \ref{fig:in3d_loss_surf}(a), (b), and (c), it can be seen that, our method enhances exploitation capability by improving sampling efficiency, while maintaining the original exploration capability from inner-iteration sampling, balancing exploration and exploitation. This ensures that adversarial samples improve sampling efficiency without sacrificing cross-model generalization, thereby increasing the attack potential of the final samples.

% ================================== Loss landscape evolving with epochs ==================================
\section{Loss Surface against Outer-Iteration}

\begin{figure}[!tb]
    % \vspace{-1em}
    \centering
    \includegraphics[width=0.45\textwidth ]{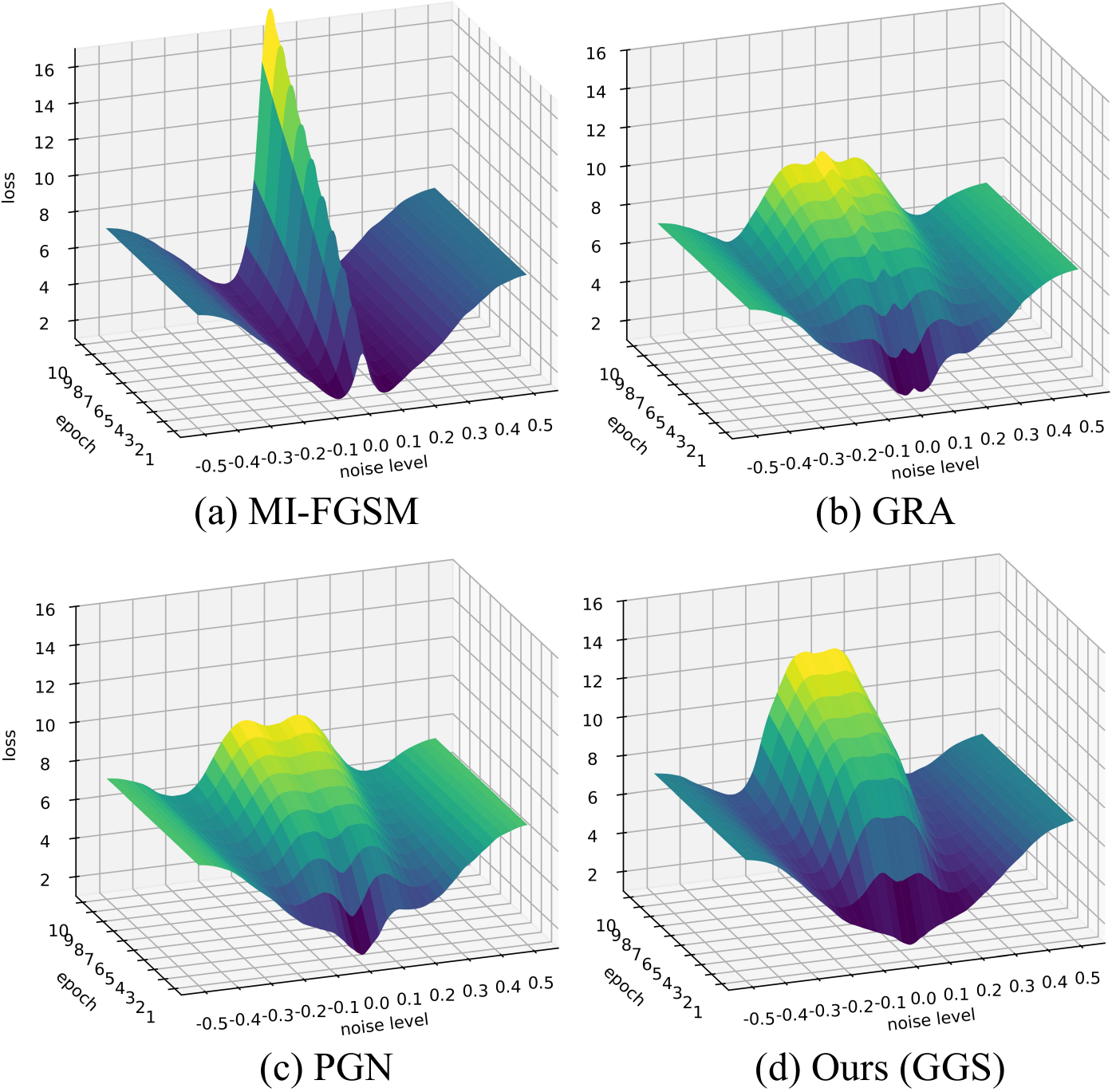} % 修改路径和尺寸
    \caption{The classification loss surfaces of adversarial examples generated by the methods (MI-FGSM \cite{dong2018mifgsm}, GRA \cite{zhu2023gra}, PGN \cite{ge2023pgn}, GGS) %changes with the increase in 
    against the number of epochs through four different methods. These adversarial examples are generated and tested on the surrogate model Res50 \cite{he2016resnet}. 
    }
    \label{fig:3d_loss_surf}
\end{figure}

\textbf{Loss Surface:} Further, we visualized the changes in the loss surface during the generation of adversarial examples using four different methods (MI-FGSM \cite{dong2018mifgsm}, GRA \cite{zhu2023gra}, PGN \cite{ge2023pgn}, and Our GGS) on the Res50 \cite{he2016resnet} model, presenting the results in a two-dimensional fashion in Fig. \ref{fig:3d_loss_surf}. 
It can be observed that, our method maintains the flatness of the loss landscape while simultaneously increasing the loss value, as the number of epochs increases.
Compared to other methods, we achieve a better balance between the magnitude of the loss values and the flatness of the loss landscape.

% ============================== More Visualizations of Loss Landscapes ==============================

\begin{figure*}[!t]
    \centering
    \includegraphics[width=0.9\textwidth ]{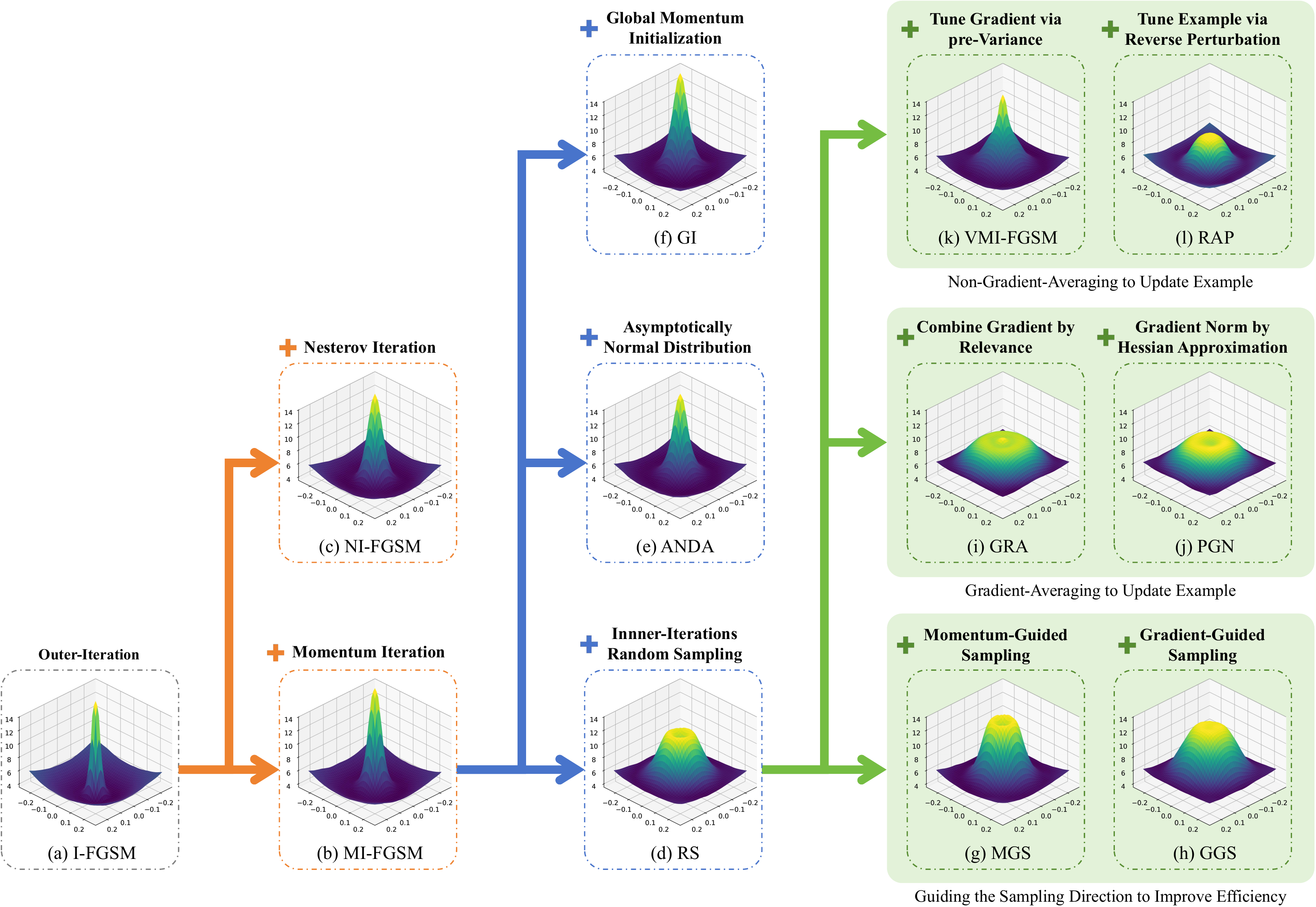} % 修改路径和尺寸
    \caption{The relationships and comparison among different methods in terms of their average loss surfaces of the 32 randomly selected adversarial examples, corresponding to all the 12 methods (I-FGSM \cite{kurakin2018ifgsm}, MI-FGSM \cite{dong2018mifgsm}, NI-FGSM \cite{lin2019nifgsm_sim}, VMI-FGSM \cite{wang2021vmifgsm}, RAP \cite{qin2022rap}, GRA \cite{zhu2023gra}, PGN \cite{ge2023pgn}, ANDA \cite{fang2024anda}, GI-FGSM \cite{wang2024gifgsm}, our GGS and its related variants RS and MGS). The x and y axes on loss surface represent the weights of two random directions, respectively, and the z-axis indicates the magnitude of the loss value obtained by testing the adversarial examples, after incorporating the combination of the corresponding two random directions like PGN \cite{ge2023pgn}.}
    \label{fig:real3d_all}
\end{figure*}

\begin{table*}
    \centering 
    % \scriptsize 
    \small
    % \vspace{-1.5em}
    \caption{The comparison of various inner-iteration methods (VMI \cite{wang2021vmifgsm}, RAP \cite{qin2022rap}, PGN \cite{ge2023pgn}, GRA \cite{zhu2023gra}, and our GGS). Out. and In. denote the numbers of outer-iterations and inner-iterations, respectively. pre-grad denote the gradient from previous inner-iteration.
 }
    % \vspace{-1.5em}
    \setlength{\tabcolsep}{3pt} % 设置列间距为 4pt
    \begin{tabular}{c|cc|ccc}
        \toprule
        & Out. & In. & \makecell{Inner-Iteration Purpose} & \makecell{Relationship} & \makecell{Outer-Iteration Purpose} \\
        \midrule
        VMI & 10 & 20  & \makecell{gradient variance} & Nested & \makecell{tune gradient via variance from previous iteration} \\
        RAP & 400 & 8  & \makecell{reverse perturbation} & \makecell{Conditional Nested} & \makecell{tune example via reverse perturbation} \\
        PGN & 10 & 20  & \makecell{gradient norm by  Hessian approximation} & Nested & \makecell{update example by momentum} \\
        GRA & 10 & 20  & \makecell{averaged gradient}  & Nested & \makecell{combine grad. by relevance \& update delta by decay} \\
        GGS & 10 & 20  & \makecell{averaged gradient lookahead by pre-grad} & Nested & \makecell{update example by momentum} \\
        \bottomrule
    \end{tabular}
    \label{tab:method_diff}
\end{table*}

\section{More Visualizations of Loss Surfaces}

To provide a more comprehensive comparison of the loss surfaces and the connections between different methods, we plotted the loss surfaces of all the nine comparison methods along with our GGS and its related variants MGS and RS methods in Fig. \ref{fig:real3d_all}. 
According to the loss surfaces shown in Fig. \ref{fig:real3d_all}, previous methods can be generally categorized into two types: those with larger local maxima and those with flatter loss surfaces. However, our GGS method achieves an excellent balance between these two characteristics by maintaining local flatness while possessing relatively large local maxima. This balance is consequently helpful to achieve strong adversarial transferability.

Additionally, to better distinguish the differences among various inner-iteration methods, we summarized these methods based on the purpose of inner and outer iterations as well as the relationship between them. This summary is presented in Table \ref{tab:method_diff}.

% ============================== Attack on Ensemble Models ==============================
\section{Attack on Ensemble Models}

As shown in Table \ref{result-table-ensemble}, our method consistently achieves higher transfer attack success rates (ASR) across multiple model architectures, compared to other existing methods, under multiple model settings, demonstrating its strong generalization capabilities. Specifically, for targeted and untargeted attacks in a multiple model ensemble setting, our method increases the average ASR by 5\%. For untargeted attacks, it achieved a score of 95.99\%, approaching a nearly complete success in attacking all the samples, revealing its strong attack capability in complex model ensemble environments.

\begin{table*}
    \caption{The average untargeted and targeted ASR ($\%$) on all the nine models, with adversarial examples generated on multiple models (Res50, Inc-v3 and ViT-B). Each data pair ($u/w$) corresponds to the performances under (untargeted/targeted) attacks. The best and second best results are labeled in \textbf{bold} and \underline{underline}, respectively.}
    
    \label{result-table-ensemble}
    \centering
    % \small
    % \tiny
    \footnotesize
    % \scriptsize
    
    \setlength{\tabcolsep}{3.6pt}
    \begin{tabular}{l|c|cccccc|ccc|c}
        \midrule \multirow{2}{*}{} 
        & Attack & \textbf{Res50} & Dense121 & \textbf{Inc-v3} & IncRes-v2 & \textbf{ViT-B} & PiT-B
        & $\text{Inc-v3}_{\text{ens3}}$ & $\text{Inc-v3}_{\text{ens4}}$ & $\text{IncRes-v2}_{\text{ens}}$ & Avg.\\

        \midrule \multirow{9}{*}{\shortstack{Res50 \\ Inc-v3 \\ ViT-B}} 
        & MI   & 99.4 / 66.7 & 73.4 / 0.5 & 100.0 / 91.2 & 64.3 / 0.3 & 98.3 / 86.8 & 51.4 / 0.1 & 44.6 / 0.0 & 46.7 / 0.0 & 36.9 / 0.0 & 68.33 / 27.29 \\
        & NI   & 99.7 / 72.4 & 79.9 / 1.0 & 100.0 / 88.1 & 71.8 / 0.9 & 98.5 / 83.6 & 51.9 / 0.3 & 48.6 / 0.1 & 51.6 / 0.1 & 40.4 / 0.0 & 71.38 / \underline{27.39} \\
        & VMI  & 99.0 / 40.7 & 84.0 / 2.3 & 100.0 / 72.4 & 81.0 / 1.8 & 97.5 / 60.9 & 66.8 / 1.1 & 63.8 / 0.2 & 65.5 / 0.5 & 57.4 / 0.4 & 79.44 / 20.30 \\
        & RAP  & 99.7 / 15.8 & 94.4 / 1.1 & 100.0 / 24.6 & 90.2 / 0.7 & 99.2 / 30.5 & 75.2 / 0.2 & 64.3 / 0.1 & 66.8 / 0.3 & 54.0 / 0.0 & 82.64 / 8.14 \\
        & GRA  & 97.3 / 26.1 & 93.3 / 4.7 & 100.0 / 58.6 & 94.1 / 5.4 & 95.7 / 48.3 & 83.0 / 4.2 & 85.4 / 1.4 & 85.6 / 2.3 & 81.7 / 1.8 & \underline{90.68} / 16.98 \\
        & PGN  & 95.8 / 13.6 & 94.5 / 4.4 & 100.0 / 44.7 & 93.9 / 5.2 & 93.0 / 25.1 & 82.5 / 3.3 & 85.1 / 1.3 & 86.1 / 1.7 & 82.1 / 1.5 & 90.33 / 11.20 \\
        & ANDA & 99.5 / 45.7 & 96.5 / 5.5 & 100.0 / 74.6 & 93.6 / 5.2 & 99.3 / 67.4 & 87.2 / 2.8 & 79.8 / 1.0 & 81.3 / 1.1 & 74.4 / 0.6 & 90.18 / 22.66 \\
        & GI   & 99.7 / 77.8 & 86.2 / 1.8 & 100.0 / 93.5 & 79.1 / 1.7 & 99.4 / 91.1 & 64.8 / 0.7 & 57.4 / 0.3 & 58.9 / 0.0 & 49.7 / 0.0 & 77.24 / 29.66 \\
        & GGS  & 99.3 / 54.1 & 98.0 / 23.8 & 100.0 / 78.8 & 98.2 / 24.9 & 99.2 / 71.4 & 94.6 / 19.7 & 91.9 / 10.6 & 93.3 / 10.3 & 89.4 / 8.1 & \textbf{95.99} / \textbf{33.52} \\

        \midrule
    \end{tabular}
\end{table*}

% ================================== Comparison of Adversarial Noise Intensity ==================================
% 不可见性指标和对比
\section{Comparison of Adversarial Perturbation}

\begin{figure}[!tb]
    \centering
    \includegraphics[width=0.48\textwidth ]{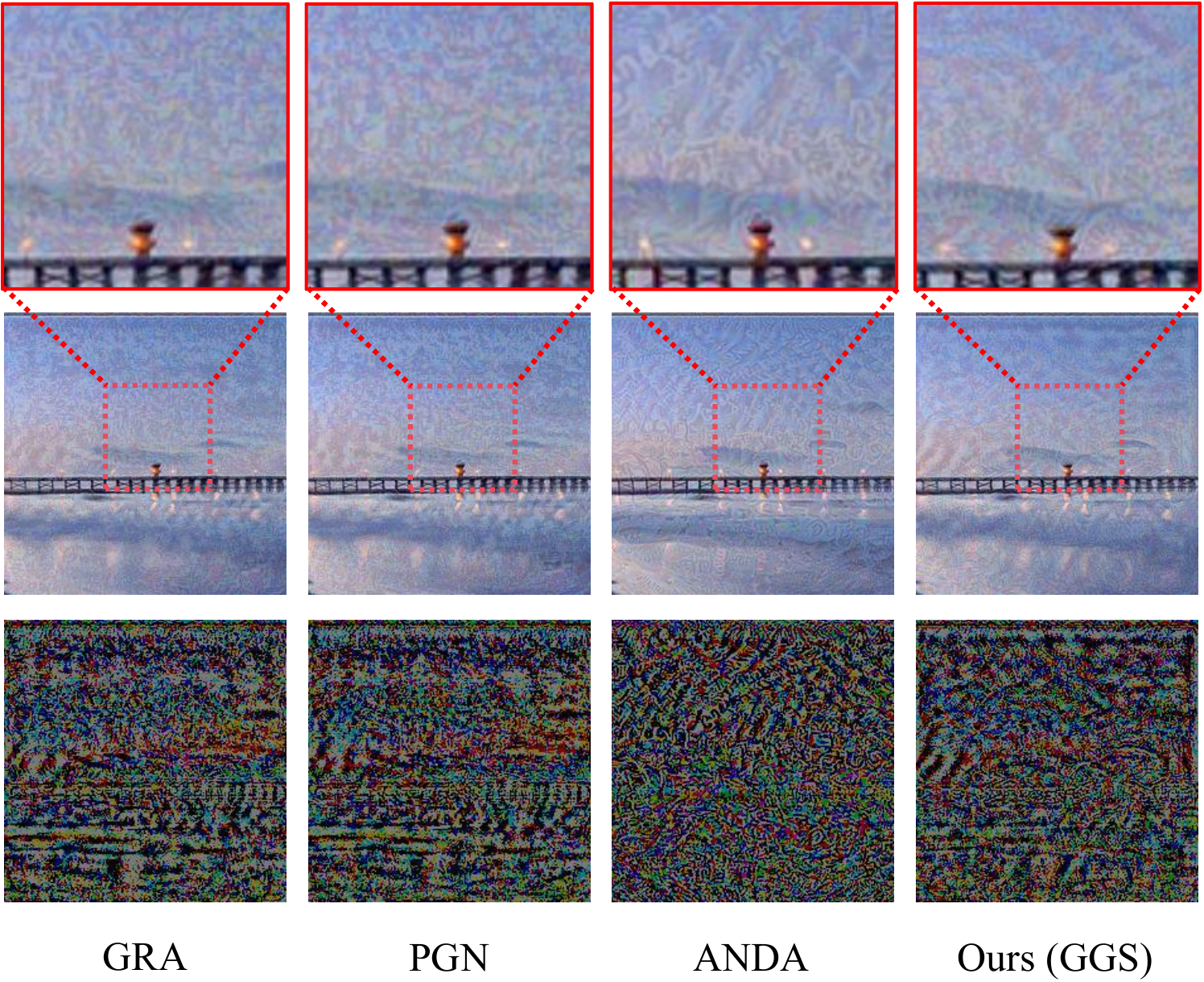} % 修改路径和尺寸
    \caption{Visualization comparison of adversarial examples generated by four advanced methods (GRA \cite{zhu2023gra}, PGN \cite{ge2023pgn}, ANDA \cite{fang2024anda}, GGS) using Res50 \cite{he2016resnet} as the surrogate model. The first row shows the locally magnified regions, the second row displays the corresponding adversarial examples, and the third row presents the adversarial noises.}
    \label{fig:impr}
\end{figure}

\begin{table}[htbp]
    \small
    \centering
    \caption{Perceptibility metrics of adversarial examples generated by different methods on Res50 \cite{he2016resnet}, including Attack Success Rate (ASR) as a measure of untargeted attack capability (average ASR across nine models), and the metrics of the imperceptibility, such as conventional L2 norm, Structural Similarity (SSIM \cite{wang2004ssim2, hameed2021ssim}) 
    , Peak Signal-to-Noise Ratio (PSNR), and average distortion of Low-Frequency Component (LF \cite{luo2022lf}).}
    \label{tab:impr}
    \setlength{\tabcolsep}{5pt}
    \begin{tabular}{c|c|cccc}
        \toprule
        Method & ASR & PSNR $\uparrow$ & SSIM $\uparrow$ & L2 $\downarrow$ & LF $\downarrow$ \\
        \midrule
        $\textbf{GRA}_{\text{ ICCV'23}}$        & 73.41 & \underline{28.31} & \underline{0.711} & \underline{17.90} & 12.93 \\
        $\textbf{PGN}_{\text{ NeurIPS'23}}$     & \underline{76.53} & 28.07 & 0.701 & 18.39 & 13.24 \\
        $\textbf{ANDA}_{\text{ CVPR'24}}$       & 64.67 & \textbf{28.60} & \textbf{0.711} & \textbf{17.32} & \textbf{12.12} \\
        \textbf{GGS}                            & \textbf{82.08} & 28.16 & 0.697 & 18.22 & \underline{12.48} \\
        \bottomrule
    \end{tabular}
\end{table}

To compare the differences in noise intensities produced by recent advanced methods, we conducted both qualitative and quantitative analyses on the adversarial examples generated by GRA \cite{zhu2023gra}, PGN \cite{ge2023pgn}, ANDA \cite{fang2024anda}, and our GGS.

\textbf{Algorithm Comparison in terms of Perturbation Metrics: }  We used four metrics, e.g. PSNR, SSIM \cite{wang2004ssim2, hameed2021ssim}, L2, and LF \cite{luo2022lf}, to evaluate the imperceptibility of adversarial perturbations relative to the clean samples. As shown in Table \ref{tab:impr}, all methods exhibit relatively similar distortion metric values. The primary reason for this is that all methods utilize a fixed step size coupled with the Fast Gradient Sign Method (FGSM \cite{Goodfellow2015fgsm}), during each adversarial example updating, which effectively limits the noise intensity within a relatively narrow range.

\textbf{Visualization of adversarial examples: }  We further visualized the adversarial examples generated by different methods and provided a local zoom-in comparison. As shown in Fig. \ref{fig:impr}, it can be observed that the adversarial noises generated by the four methods do not exhibit significant differences at the scale of the entire image. However, upon closer inspection, PGN \cite{ge2023pgn}, GRA \cite{zhu2023gra} and our GGS generate perturbations with lower regularity in shape (fewer continuous groove-like perturbations) 
% and higher average ASR compared to other methods
. This may be a manifestation of high transferability in the noise patterns, and we hope that future work can uncover the relationship between noise morphology and transferability.

\section{GGS vs. NCS with different random seeds}
\label{exp:visual}

Table \ref{tab:add_method} shows the mean and avg. range from 10 independent runs with random seeds. GGS surpassed NCS \cite{qiu2024ncs} using Res50 and ViT-B as surrogate models.

\begin{table}[htbp]
    \centering
    \scriptsize 
    % \vspace{-1.5em}
    \caption{Average untargeted ASR ($\%$) of NCS and GGS.}
    % \vspace{-1.5em}
    \label{tab:add_method}
    
    % \begin{tabular}{c|c|cccccc|c}
    \begin{tabular}{@{\hspace{0.1em}}l@{\hspace{0.2em}}|
    @{\hspace{0.3em}}c@{\hspace{0.3em}}|
    @{\hspace{0.3em}}c@{\hspace{0.5em}}c@{\hspace{0.4em}}c@{\hspace{0.4em}}c@{\hspace{0.4em}}c@{\hspace{0.7em}}c
    @{\hspace{0.5em}}|
    @{\hspace{0.4em}}c@{\hspace{0.1em}}}
    
        \toprule 
        % \multirow{2}{*}{}
        & Attack & \textbf{Res50} & Dense121 & \textbf{Inc-v3} & IncRes-v2 & \textbf{ViT-B} & PiT-B & Avg.\\
        
        \midrule \multirow{2}{*}{\makecell{Res50}}
        & NCS     & 96.16 & 87.00 & 80.20 & 73.46 & 53.58 & 67.74 & 76.36  \textcolor{gray}{$\pm$ 0.27} \\
        & GGS     & 99.30 & 95.78 & 89.46 & 85.92 & 60.14 & 80.18 & \textbf{85.13} \textcolor{gray}{$\pm$ 0.18} \\
        \midrule \multirow{2}{*}{\makecell{Inc-v3}}
        & NCS    & 69.38 & 84.52 & 99.76 & 91.20 & 32.54 & 45.10 & \textbf{70.42} \textcolor{gray}{$\pm$ 0.30} \\
        & GGS    & 68.42 & 86.72 & 100.00 & 95.72 & 27.64 & 39.16 & 69.61 \textcolor{gray}{$\pm$ 0.74} \\
        \midrule \multirow{2}{*}{\makecell{ViT-B}}
        & NCS     & 72.18 & 82.00 & 79.34 & 72.72 & 98.56 & 85.32 & 81.69 \textcolor{gray}{$\pm$ 0.15} \\
        & GGS     & 81.26 & 90.12 & 86.68 & 81.84 & 100.0 & 92.50 & \textbf{88.73} \textcolor{gray}{$\pm$ 0.50} \\

        \bottomrule
    \end{tabular}
    % \vspace{-2em}
\end{table}

\section{Inner-Iterations vs. Efficiency analysis}

\begin{figure}[H]
    \vspace{-0.9em}
    \centering
    \includegraphics[width=0.48\textwidth ]{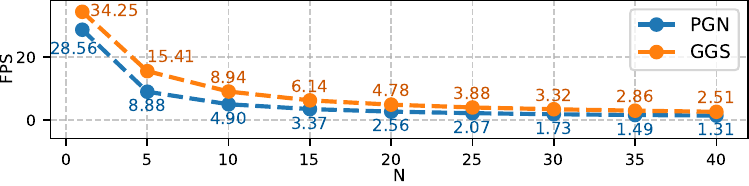} % 修改路径和尺寸
    \vspace{-2.3em}
    \caption{The FPS of PGN and GGS on Res50.}
    \vspace{-1.5em}
    \label{fig:fps}
\end{figure}

Fig. \ref{fig:fps} shows PGN and Our method’s time consumption scales with inner iterations, while GGS maintaining 2.5 FPS compare with PGN's 1.3 FPS at 40 iterations. N=20 was set to ensure usability and fair comparison with other methods.

% ================================== FPS and GPU Memory ==================================

\section{Inference Speeds and GPU Memory Usage}

\begin{table}[htbp]
\small 
\centering
    \caption{Comparison of inference speeds in terms of Frames Per Second (FPS) and GPU Memory Usage (GB) of different methods. The official default batch size (bs) of ANDA \cite{fang2024anda} is set to 1, RAP is set to 16 due to memory constraints, while for other methods, the default batch size (bs) is set to 64 for Res50 \cite{he2016resnet} and Inc-v3 \cite{szegedy2016incv3}, and 32 for ViT-B \cite{dosovitskiy2020vit}. 
    The MI-FGSM \cite{dong2018mifgsm}, NI-FGSM \cite{lin2019nifgsm_sim} and GI-FGSM \cite{wang2024gifgsm} do not include an inner-iteration process, thus resulting in a faster inference speed.
    }
    \label{tab:speed}
    \setlength{\tabcolsep}{3pt}
    \begin{tabular}{c|ccc|ccc}
        \toprule
        \multirow{2}{*}{Method} & \multicolumn{3}{c|}{{FPS $\uparrow$}} & \multicolumn{3}{c}{{GB $\downarrow$}} \\
         & Res50 & Inc-v3 & ViT-B & Res50 & Inc-v3 & ViT-B \\
        \midrule
        $\textbf{MI}_{\text{ CVPR'18 \cite{dong2018mifgsm}}}$     & 43.5 & 52.6 & 27.0 & 8.37 & 5.96 & 5.47\\
        $\textbf{NI}_{\text{ ICLR'20 \cite{lin2019nifgsm_sim}}}$  & 47.6 & 41.7 & 26.3 & 8.36 & 5.96 & 5.47\\
        $\textbf{GI}_{\text{ ESWA'24 \cite{wang2024gifgsm}}}$     & 31.3 & 27.8 & 18.2 & 8.38 & 6.02 & 5.47\\
        \midrule
        $\textbf{VMI}_{\text{ CVPR'21 \cite{wang2021vmifgsm}}}$ & 4.7 & 6.4 & \underline{1.8} & \underline{8.60} & \underline{6.14} & 5.56 \\
        $\textbf{RAP}_{\text{ NeurIPS'22 \cite{qin2022rap}}}$ & 0.25 & 0.15 & 0.13 & 21.3 & 20.5 & 21.5 \\
        $\textbf{GRA}_{\text{ ICCV'23 \cite{zhu2023gra}}}$ & \underline{4.7} & \underline{6.4} & 1.8 & 8.76 & 6.35 & 5.55\\
        $\textbf{PGN}_{\text{ NeurIPS'23 \cite{ge2023pgn}}}$ & 2.5 & 3.6 & 1.0 & 8.93 & 6.26 & 5.55\\
        $\textbf{ANDA}_{\text{ CVPR'24 \cite{fang2024anda}}}$& 3.6 & 2.4 & 1.4 & \textbf{3.23} & \textbf{2.42} & \textbf{4.57}\\
        
        \textbf{GGS} & \textbf{4.9} & \textbf{6.6} & \textbf{1.9} & 8.78 & 6.24 & \underline{5.53}\\
        \bottomrule
    \end{tabular}
\end{table}

To compare the differences in efficiency and overhead of various methods for generating adversarial examples, we conducted a detailed evaluation of the inference speed and GPU memory usage of all nine methods, as shown in Table \ref{tab:speed}. Since the MI-FGSM \cite{dong2018mifgsm}, NI-FGSM \cite{lin2019nifgsm_sim} and GI-FGSM \cite{wang2024gifgsm} methods do not involve an inner-iteration process, we focused primarily on the remaining six methods that include an inner-iteration process for a fair comparison. It can be observed that our GGS method exhibits the fastest inference speed, and the GPU memory usage does not obviously increase. Notably, due to algorithmic limitations, ANDA \cite{fang2024anda} can be only configured with a batch size of 1, resulting in a reduced GPU memory usage.

% ============================== Sensitivity Analysis of Hyperparameters ==============================
\section{Sensitivity Analysis of Hyperparameters}

\begin{figure}[!tb]
    % \vspace{-1em}
    \centering
    \includegraphics[width=0.48\textwidth ]{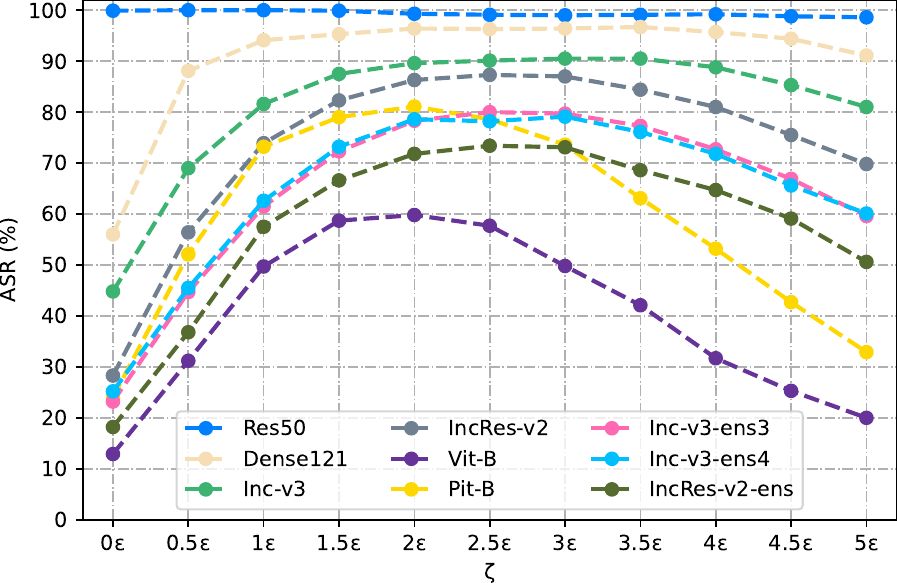} 
    \caption{Untargeted attack success rate (ASR, \%) on different models (Res50 \cite{he2016resnet}, Dense121 \cite{huang2017densenet}, Inc-v3 \cite{szegedy2016incv3}, IncRes-v2 \cite{szegedy2017incresv2}, ViT-B \cite{dosovitskiy2020vit}, PiT-B \cite{heo2021pit}, $\text{Inc-v3}_{\text{ens3}}$ \cite{tramer2017incens}, $\text{Inc-v3}_{\text{ens4}}$ \cite{tramer2017incens}, $\text{IncRes-v2}_{\text{ens}}$ \cite{tramer2017incens}) with different values of $\zeta$, with Res50 as a surrogate model. The maximum perturbation $\epsilon =16/255$.}
    \label{zeta_line}
    % \vspace{-1em}
\end{figure}

\textbf{Sensitivity Analysis of $\zeta$.}
In Fig. \ref{zeta_line}, we analyze the impact of the upper bound neighborhood size, determined by parameter $\zeta$, on the attack success rate (ASR) in black-box settings. The experiment employs uniform sampling to reduce bias from uneven sample distribution. As $\zeta$ gradually increases to $2.0 \times \epsilon$, the transferability performance of normally trained models is improved and reaches its peak, while the transferability of adversarially trained models continues to rise. However, when $\zeta$ exceeds $3.5 \times \epsilon$, the adversarial transferability of the eight black-box models begins to decline. To achieve a balance between normally and adversarially trained models in transferability, we select $\zeta = 2.0 \times \epsilon$ for the experiments.

% ================================== Sensitivity of N ==================================

\textbf{Sensitivity Analysis of $N$.}
\begin{figure}[!tb]
    % \vspace{-1em}
    \centering
    \includegraphics[width=0.45\textwidth ]{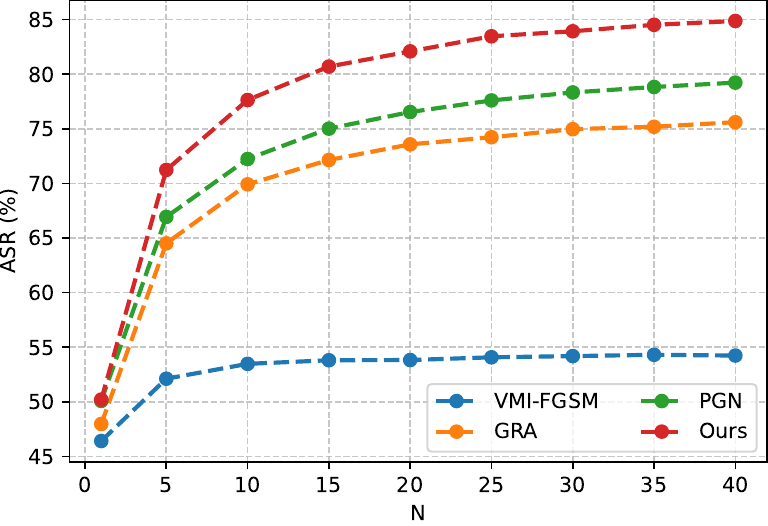} 
    \caption{The average attack success rate on nine models with different $N$ (inner-iterations numbers) across different methods (VMI-FGSM \cite{wang2021vmifgsm}, GRA \cite{zhu2023gra}, PGN \cite{ge2023pgn}, our GGS).}
    \label{iter_asr}
    % \vspace{-1em}
\end{figure}
We further conducted experimental evaluations to assess the impact of varying the number of inner-iterations $N$ on different methods. The results are presented in Fig. \ref{iter_asr}. It can be observed that, the attack success rate for all the methods has been largely improved when the number of inner-iterations is within 15. As the number of inner-iterations increases, the effectiveness of GRA \cite{zhu2023gra}, PGN \cite{ge2023pgn}, and our GGS method can be further enhanced. Considering both time overhead and effectiveness, the selection of 20 inner-iterations (N=20) provides a balanced benefit.

% ============================== Experimental Settings ==============================

\section{Detailed Hyperparameter Settings}
% \textbf{Hyperparameters:} 
For all methods, we set the maximum perturbation of the parameter $\epsilon = 16/255$. For all methods, except RAP, we set the number of outer-iteration is $T = 10$, the step size is $\alpha = \epsilon/T$.
For VMI-FGSM \cite{wang2021vmifgsm}, we configure the number of sampled examples as $N = 20$ and set the upper bound of neighborhood size to $\beta = 1.5 \times \epsilon$;
For RAP, the step size is $\alpha = 2.0/255$, with the number of iterations $K = 400$, an inner-iteration number $T = 8$, and a late-start $K_{LS} = 100$. The size of neighborhoods is $\epsilon_n = 16/255$. 
For GRA \cite{zhu2023gra}, the number of sampled examples is $N = 20$, the sample quantity is $m = 20$, the upper bound factor of sample range is $\beta=3.5$ and the attenuation factor is $\eta=0.94$;
For PGN \cite{ge2023pgn}, the number of sampled examples is $N = 20$, the balanced coefficient is $\delta = 0.5$, and the upper bound factor of sample range is $\zeta = 3.0 \times \epsilon$.
For ANDA \cite{fang2024anda}, we strictly adhere to the hyperparameter settings specified in its official code repository.
For GI-FGSM \cite{wang2024gifgsm}, the pre-convergence iterations $P = 5$, and the global search factor $S = 10$.

% ========================= Details combined with different methods ============================

\section{Detailed Results when our GGS is Combined with Different Methods}

Due to space limit in the main text, we only present the average attack success rates of our GGS combined with different methods. More detailed results, including targeted and untargeted attack across all nine models, are provided in Table \ref{ours-plus-table}.

\begin{table*}
    \caption{The untargeted and targeted ASR ($\%$) on all nine models with the adversarial examples generated on ResNet50, when combined our GGS with input transformation-based methods (DIM\cite{xie2019dim}, TIM\cite{dong2019tim}, SIM\cite{lin2019nifgsm_sim}, Admix\cite{wang2021admix}, SSM\cite{long2022ssm}). Each data pair ($u$/$w$) corresponds to the performances under (original method / variant combined with our GGS).
    }
    \label{ours-plus-table}
    \centering
    % \small
    % \tiny
    \footnotesize
    % \scriptsize
    \begin{tabular}{@{\hspace{0.5em}}c@{\hspace{0.5em}}|
    @{\hspace{0.6em}}c@{\hspace{1.1em}}c@{\hspace{1.1em}}c@{\hspace{1.1em}}c
    @{\hspace{1.1em}}c@{\hspace{1.1em}}c@{\hspace{0.6em}}|
    @{\hspace{0.6em}}c@{\hspace{1.1em}}c@{\hspace{0.5em}}c@{\hspace{0.2em}}|
    @{\hspace{0.6em}}c@{\hspace{0.5em}}}
        \toprule
        Method & Res50 & Dense121 & Inc-v3 & IncRes-v2 & Vit-B & PiT-B
        & $\text{Inc-v3}_{\text{ens3}}$ & $\text{Inc-v3}_{\text{ens4}}$ & $\text{IncRes-v2}_{\text{ens}}$ & Avg.\\
        \midrule 
            \multicolumn{11}{c}{{untargeted attack}} \\
        \midrule
        DIM / +ours     & 97.6 / 98.6 & 72.8 / 96.3 & 60.3 / 94.3 & 50.8 / 93.7 & 25.6 / 75.3 & 41.5 / 88.2 & 39.3 / 89.6 & 40.9 / 88.6 & 32.9 / 86.6 & 51.30 / \textbf{90.13} \\
        TIM / +ours     & 99.9 / 99.3 & 60.0 / 96.0 & 49.9 / 90.7 & 33.8 / 87.5 & 15.7 / 63.6 & 26.6 / 79.9 & 28.1 / 84.3 & 30.1 / 83.1 & 23.3 / 80.1 & 40.82 / \textbf{84.94} \\
        SIM / +ours     & 100.0 / 99.6 & 71.8 / 98.8 & 56.2 / 95.4 & 40.6 / 94.2 & 22.0 / 71.9 & 36.2 / 89.3 & 32.3 / 89.5 & 35.2 / 88.4 & 26.9 / 84.0 & 46.80 / \textbf{90.12} \\
        Admix / +ours   & 100.0 / 99.7 & 82.7 / 99.2 & 65.4 / 95.9 & 51.4 / 95.0 & 28.5 / 73.6 & 46.8 / 89.0 & 41.9 / 89.5 & 43.1 / 88.3 & 33.2 / 85.5 & 54.78 / \textbf{85.50} \\
        SSM / +ours     & 97.9 / 98.9 & 89.7 / 96.1 & 81.4 / 92.2 & 77.6 / 88.8 & 50.1 / 47.4 & 67.3 / 70.2 & 70.0 / 80.5 & 69.3 / 79.3 & 64.7 / 73.3 & 74.22 / \textbf{80.74} \\
        % \midrule
        GRA / +ours     & 96.9 / 99.2 & 88.6 / 96.8 & 81.8 / 91.0 & 75.8 / 85.7 & 45.3 / 44.0 & 62.6 / 66.4 & 71.5 / 77.5 & 70.9 / 76.6 & 67.3 / 71.0 & 73.41 / \textbf{78.69} \\
        PGN / +ours     & 98.6 / 99.8 & 91.3 / 97.4 & 85.0 / 91.3 & 78.5 / 88.8 & 49.7 / 55.2 & 67.8 / 78.4 & 74.9 / 81.9 & 72.9 / 80.7 & 70.1 / 75.6 & 76.53 / \textbf{83.23}  \\
        \midrule
            \multicolumn{11}{c}{{targeted attack}} \\
        \midrule
        DIM / +ours     & 76.1 / 83.3 & 1.1 / 26.7 & 0.1 / 9.9 & 0.2 / 13.3 & 0.0 / 6.7 & 0.1 / 11.1 & 0.0 / 7.7 & 0.0 / 7.5 & 0.1 / 8.1 & 8.63 / \textbf{19.37} \\
        TIM / +ours     & 98.9 / 89.7 & 0.2 / 28.9 & 0.0 / 7.9 & 0.1 / 9.7 & 0.0 / 4.1 & 0.0 / 8.6 & 0.0 / 6.1 & 0.0 / 5.5 & 0.1 / 5.8 & 11.03 / \textbf{18.48} \\
        SIM / +ours     & 99.1 / 95.8 & 0.9 / 52.5 & 0.1 / 17.6 & 0.0 / 21.8 & 0.0 / 10.9 & 0.1 / 22.1 & 0.0 / 11.6 & 0.0 / 13.4 & 0.1 / 10.3 & 11.14 / \textbf{28.44} \\
        Admix / +ours   & 97.3 / 96.6 & 3.8 / 57.6 & 0.0 / 19.0 & 0.1 / 23.5 & 0.0 / 11.9 & 0.2 / 24.9 & 0.0 / 15.2 & 0.1 / 14.0 & 0.0 / 11.8 & 11.28 / \textbf{30.50} \\
        SSM / +ours     & 63.5 / 69.1 & 6.2 / 18.1 & 1.2 / 5.3 & 1.8 / 5.6 & 0.3 / 1.3 & 0.6 / 4.0 & 0.5 / 3.5 & 0.8 / 3.0 & 0.7 / 3.0 & 8.40 / \textbf{12.54} \\
        % \midrule
        GRA / +ours     & 67.7 / 83.2 & 3.9 / 19.9 & 0.9 / 4.5 & 1.7 / 5.6 & 0.2 / 0.9 & 0.6 / 3.0 & 0.7 / 1.4 & 1.1 / 2.4 & 0.7 / 1.9 & 8.61 / \textbf{13.64}  \\
        PGN / +ours     & 49.5 / 73.3 & 4.7 / 19.2 & 1.4 / 5.1 & 1.7 / 5.7 & 0.4 / 1.5 & 1.0 / 3.2 & 0.6 / 2.2 & 1.1 / 2.5 & 1.2 / 2.4 & 6.84 / \textbf{12.79}  \\
        \bottomrule
    \end{tabular}
\end{table*}

% ============================== Attacking Multimodal Large Language Models ==============================
\section{Details of the attack on MLLMs}

Specifically, (1) 4o and mini represent gpt-4o-2024-08-06 and gpt-4o-mini-2024-07-18, respectively; (2) pro and flash represent gemini-1.5-pro-002 and gemini-1.5-flash-002, respectively; (3) sonnet represent claude-3-5-sonnet-20240620. 
For the system prompt, to better study the image content understanding capabilities of MLLMs while simplifying the experimental process, we converted the multi-classification problem into a binary classification one. Specifically, we input both the image label name and the image itself into the MLLMs, prompting the MLLMs to determine whether the image content belongs to the corresponding label, and provide a confidence level for its judgment. For all the five MLLMs, we uniformly applied the following system prompts:

\begin{lstlisting}[language=]
You are a strict image classification expert system. Execute the following instructions:

1. Determine whether the main content of the image belongs to the given label through visual analysis only
2. Strictly return results according to the following JSON schema:
   {"match": 1 or 0, "confidence": integer between 1-10}
3. Judgment rules:
   - "match": 1 if completely matching the label, 0 if not matching
   - "confidence": Objectively assess the credibility of your judgment based on image clarity, content features, etc.

Key constraints:
    - Prohibited from outputting any text other than JSON
    - Ensure output can be directly parsed by standard JSON parsers
    - Output must be valid JSON format
    - Please do not return empty content, you must give match and confidence, even if the picture is blank

Example output: 
    {"match": 1, "confidence": 8}
\end{lstlisting}

\end{document}